\documentclass{clv3}
\usepackage{hyperref}
\usepackage{xcolor}
\definecolor{darkblue}{rgb}{0, 0, 0.5}
\hypersetup{colorlinks=true,citecolor=darkblue, linkcolor=darkblue, urlcolor=darkblue}
\usepackage[english]{babel}
\usepackage{graphicx}
\usepackage{caption}
\usepackage{subcaption}
\usepackage{booktabs}
\usepackage{multirow}
\usepackage{array}
\usepackage[utf8]{inputenc}
\usepackage{amsmath,algorithm}
\DeclareMathOperator*{\argmax}{argmax}
\usepackage{algorithmic}
\newcolumntype{+}{>{\global\let\currentrowstyle\relax}}
\newcolumntype{^}{>{\currentrowstyle}}
\newcommand{\rowstyle}[1]{\gdef\currentrowstyle{#1}%
  #1\ignorespaces
}
\newcommand{\ssymbol}[1]{^{\@fnsymbol{#1}}}

\begin{document}

\title{Stochastic Natural Language Generation Using Dependency Information}

\author{Elham Seifossadat\thanks{Department of Computer Engineering, Sharif University
of Technology, Tehran 11365-8639, Iran. E-mail: seifossadat@ce.sharif.edu.}}
\affil{Sharif University of Technology}

\author{Hossein Sameti \thanks{Department of Computer Engineering, Sharif University
of Technology, Tehran 11365-8639, Iran. E-mail: sameti@sharif.edu.}}
\affil{Sharif University of Technology}

\maketitle

\begin{abstract}
This article presents a stochastic corpus-based model for generating natural language text. Our model first encodes dependency relations from training data through a feature set, then concatenates these features to produce a new dependency tree for a given meaning representation, and finally generates a natural language utterance from the produced dependency tree. We test our model on nine domains from tabular, dialogue act and RDF format. Our model outperforms the corpus-based state-of-the-art methods trained on tabular datasets and also achieves comparable results with neural network-based approaches trained on dialogue act, E2E and WebNLG datasets for BLEU and ERR evaluation metrics. Also, by reporting Human Evaluation results, we show that our model produces high-quality utterances in aspects of informativeness and naturalness as well as quality.
\end{abstract}

\section{Introduction}

\begin{figure}
    \begin{subfigure}[b]{0.5\textwidth}
        \includegraphics[width=\textwidth]{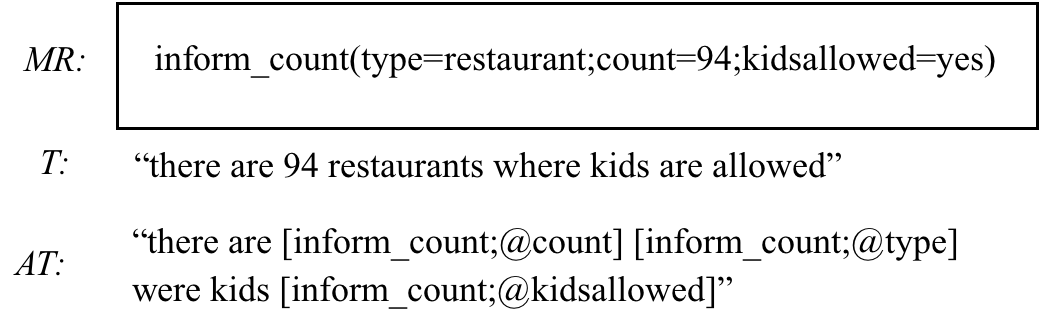} 
        \caption{Restaurant}
        \label{fig:a}
    \end{subfigure}
    \begin{subfigure}[b]{0.5\textwidth}
         \includegraphics[width=\textwidth]{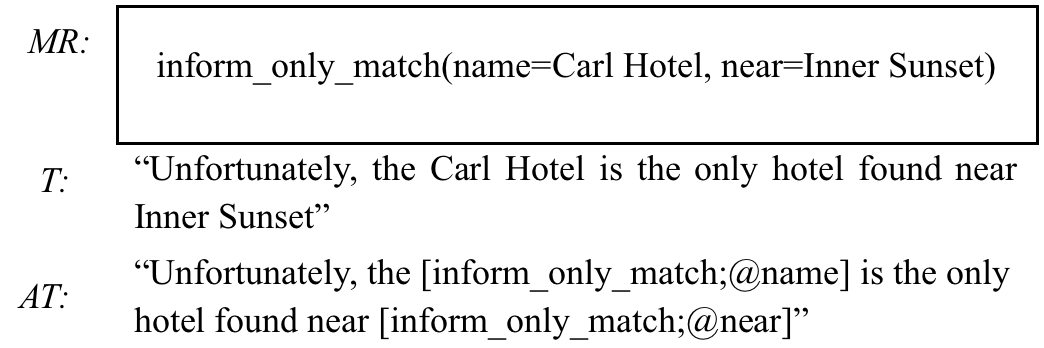}
        \caption{Hotel}
        \label{fig:b}
    \end{subfigure}
        \begin{subfigure}[b]{0.5\textwidth}
        \includegraphics[width=\textwidth]{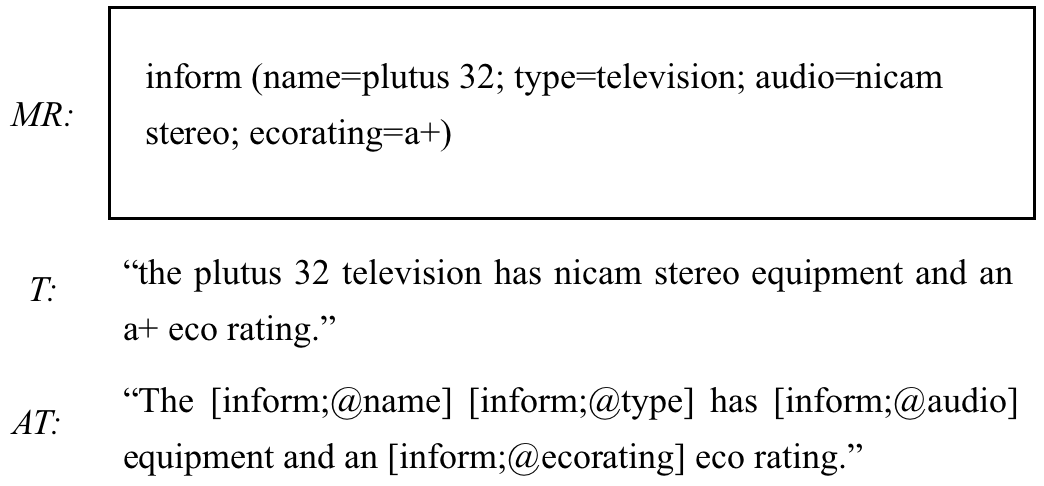} 
        \caption{TV}
        \label{fig:c}
    \end{subfigure}
    \begin{subfigure}[b]{0.5\textwidth}
         \includegraphics[width=\textwidth]{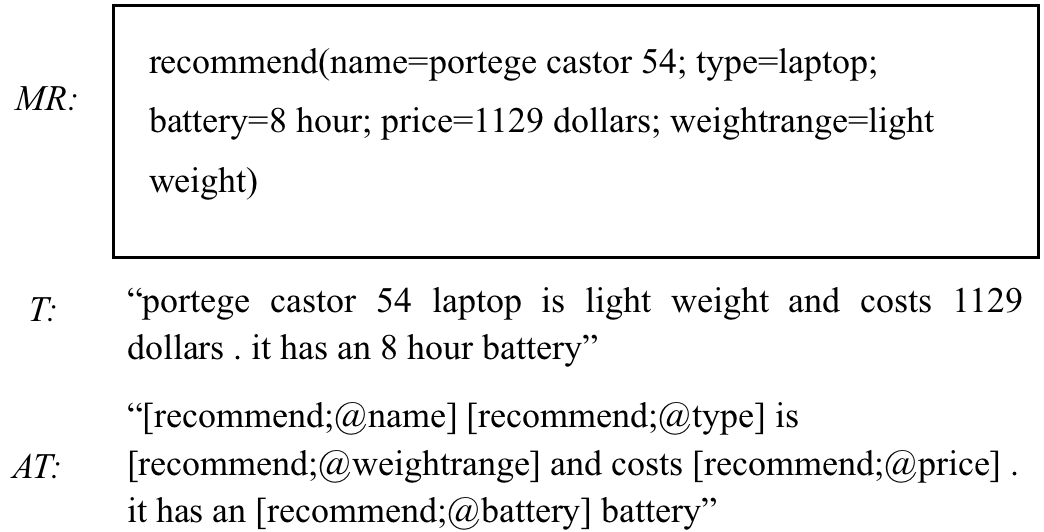}
        \caption{Laptop}
        \label{fig:d}
    \end{subfigure}
        \begin{subfigure}[b]{0.5\textwidth}
        \includegraphics[width=\textwidth]{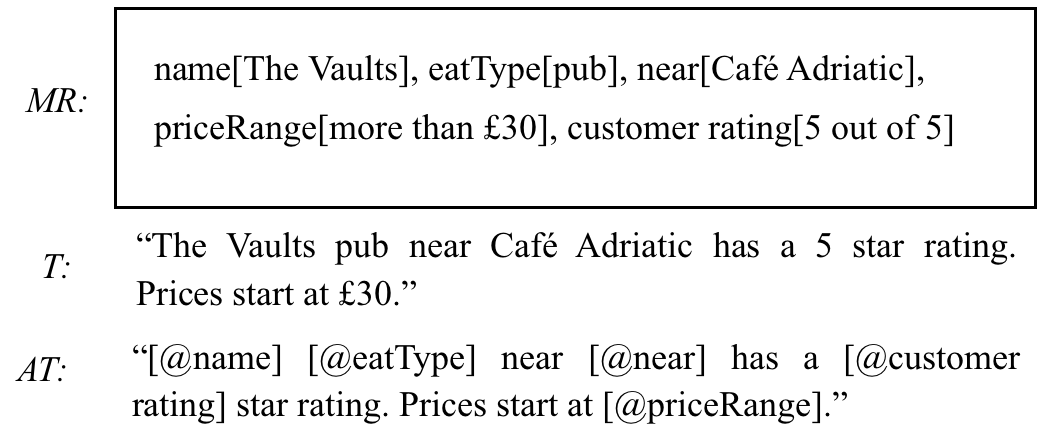} 
        \caption{E2E}
        \label{fig:e}
    \end{subfigure}
    \begin{subfigure}[b]{0.5\textwidth}
        \includegraphics[width=\textwidth]{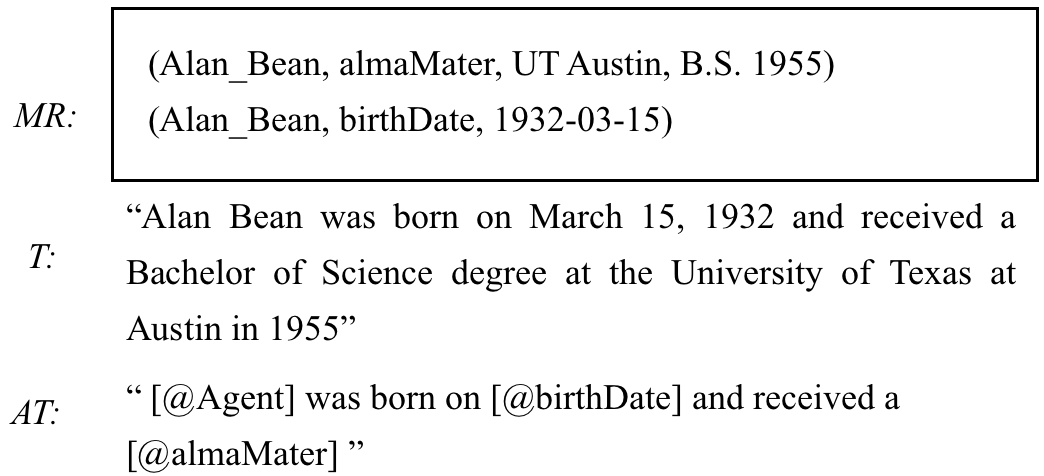} 
        \caption{WebNLG}
        \label{fig:f}
    \end{subfigure}
    \begin{subfigure}[b]{0.5\textwidth}
        \includegraphics[width=\textwidth]{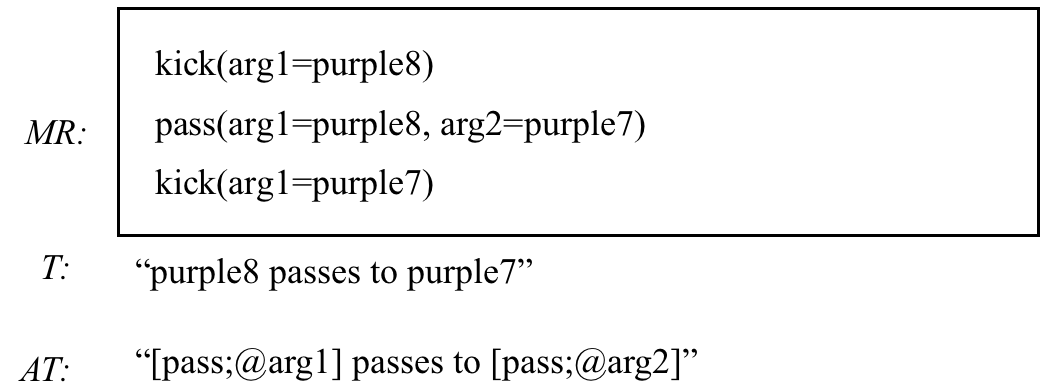} 
        \caption{RoboCup}
        \label{fig:g}
    \end{subfigure}    
    \begin{subfigure}[b]{0.5\textwidth}
        \includegraphics[width=\textwidth]{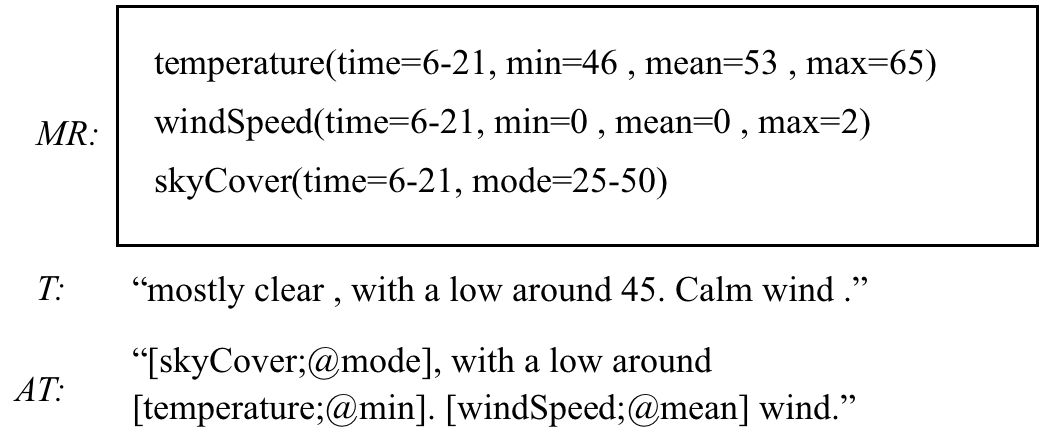} 
        \caption{WEATHERGOV}
        \label{fig:h}
    \end{subfigure}    
    \begin{subfigure}[b]{0.5\textwidth}
        \includegraphics[width=\textwidth]{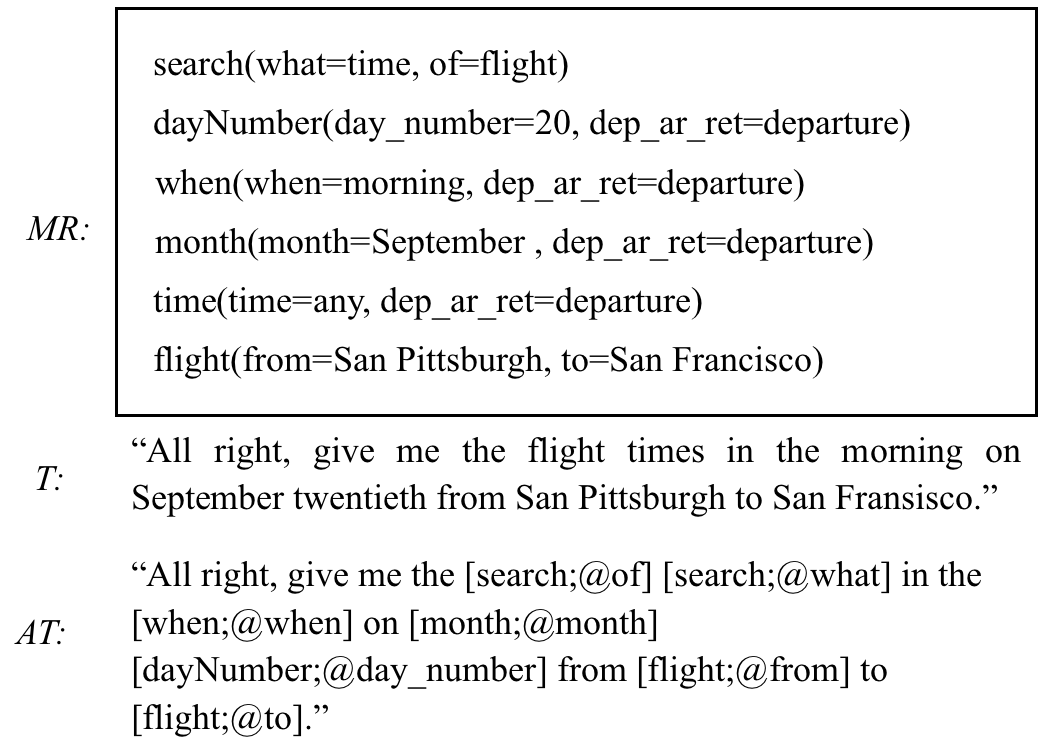} 
        \caption{Atis}
        \label{fig:i}
    \end{subfigure}          
\captionsetup{%
    font={small}, 
    format=plain,  
    singlelinecheck=true,      
    labelsep=newline,
    labelfont=bf,      
}    
\caption{ Sample meaning representation (MR), text (T) and an aligned text (AT) chosen from (a) Restaurant, (b) Hotel, (c) TV, (d) Laptop, (e) E2E, (f) WebNLG, (g) RoboCup, (h) WEATHERGOV and (i) Atis datasets.}
\label{fig: Figure1}
\end{figure}

\textit{Natural language generation} (NLG) is the task of generating a natural-language text from structured, formal and abstract meaning representation (MR) \cite{Reiter}. It is used in the generation of short data summaries, question answering, Machine Translation (MT), Spoken Dialogue Systems (SDSs), selective generation systems and search engines. Examples of meaning representations and their corresponding text are shown in Figure \ref{fig: Figure1}. For a given MR, a generator must produce a readable, fluent and adequate sentence that express all the required information contained in the MR. Moreover, a language generator should be able to produce diverse sentences for the same concept. In traditional language generator architectures, this task is often formulated as two sub-problems: \textit{sentence planning}, which decides on the overall sentence structure, and \textit{surface realization}, which converts the sentence structure into final utterance. 

Earlier approaches on statistical NLG were typically combinations of a handcrafted component and a statistical training method \cite{Knight,Langkilde,Ringger}. Although these systems produce high-quality text, they are very expensive to build, have limited linguistic coverage and variety, nor they can be used for other domains. As a solution, then, the corpus-based approaches emerged which were dependent on statistical learning of correspondences between MR and text in training data  \cite{Duboue,BarzilayL,BarzilayM,Soricut,Wong,Belz,Liang,LuS,Kim,Angeli,LuN,KonstasA,KonstasB,Gyawali}. These systems are fairly inexpensive, more adaptable and rely on having enough data for the given domain; however, more prone to errors and the output text may not be coherent and fluent as there are fewer constraints on the generated text. The most successful NLG systems use recurrent neural networks (RNNs) paired with an encoder-decoder system architecture  \cite{WenA,WenB,WenC,WenD,Mei,Dusek,Lebret,TranA,TranB,Liu,Sha}. These NLG models, however, require a greater amount of data for training. Also the sentences produced by them still have grammatical and semantic defects as well as limited naturalness  \cite{Nayak}. 

In this paper, we present a new, conceptually simple stochastic corpus-based algorithm for generating natural language sentences. The proposed algorithm focuses on encoding dependency relationships between words of unaligned training utterances to improve the grammatical structure as well as the naturalness of the generated sentences. For a given test MR, our algorithm produces the most probable dependency tree by using the learned dependency information, so that it contains the meaning concepts in the MR. We extensively conducted experiments on nine different NLG domains: Atis \cite{KonstasA}, WEATHERGOV \cite{Liang} and RoboCup \cite{Chen} in tabular format, Hotel, Restaurant, TV and Laptop information \cite{WenA,WenD} in dialogue act format and also datasets of E2E and WebNLG challenges \cite{Novikova,Colin}. We also found that the proposed method significantly outperformed the state-of-the-art methods \cite{Kim,Angeli,KonstasA,KonstasB} trained on tabular datasets and achieved competitive performance comparing to the neural network-based methods \cite{WenA,WenB,WenC,WenD,Mei,TranA,TranB,Gardent,DusekE2E} trained on dialogue act and NLG challanges datasets according the BLEU and ERR scores. In order to assess the subjective performance of our system, we used the Human Evaluation test. The results show that our approach can produce high-quality and fluent utterances. In summary, the main contributions of our work are: 1) We present a new way of constructing the structure of sentences that empirically shows improved performance compared with the state-of-the-art systems, and, 2) Our approach is independent of NLG task types, i.e., Spoken Dialogue Systems and selective generation system. While other reported works have only been tested on one of the above tasks. 

The rest of this article is organized as follows: related work is reviewed in Section 2. The proposed algorithm is presented in Section 3. Section 4 describes datasets, experimental setups, evaluation metrics and results along with their analysis. We conclude with a brief summary in Section 5. 

\section{Related Work}

Our work is situated within the broader class of stochastic data-driven approaches for content planning and surface realization. Many approaches have been proposed to learn the individual modules. For content planning module, trending approaches are aligning meaning representations and sentences as a classification problem \cite{Duboue,BarzilayM}, using hidden Markov models \cite{BarzilayL} or a hierarchical semi-Markov method \cite{Liang}. Surface realization is often treated as a problem of producing text according to a given grammar. \namecite{Soricut} propose a language generation system that uses the WIDL-representation, a formalism used to compactly represent probability distributions over finite sets of strings. \namecite{Wong}, \namecite{Belz} and \namecite{LuN} use context-free grammars to generate natural language sentences from formal meaning representations. Other effective approaches include the use of tree conditional random fields \cite{LuS}, hybrid alignment tree \cite{Kim}, template extraction within a log-linear framework \cite{Angeli}, and tree-adjoining grammar \cite{Gyawali}. Recent works combine content selection and surface realization in a unified framework \cite{KonstasA,KonstasB}.

Due to the recent successes in Deep Learning, researchers started to use end-to-end systems to jointly model the traditionally separated tasks of content planning and surface realization in one system. In weather forecasting and sports domain, \namecite{Mei} proposed an Encoder-attention-Decoder model that used both local and global attention for content selection and generating corresponding sentences. \namecite{WenA,WenD} created a dataset containing dialogue acts of four different domains: finding a restaurant, finding a hotel, buying a laptop and buying a television. Then they proposed heuristic gated LSTMs with CNN ranking method \cite{WenA}, Encoder-decoder LSTMs with attention architecture \cite{WenC} and semantically-conditioned LSTMs with backward LSTMs ranking method \cite{WenB,WenD} for encoding dialogue acts and generating sentences word by word in the decoding step. To improve \namecite{WenC} results, \namecite{TranA} suggested using an attention mechanism for representing dialogue acts and then refining the input token based on this representation. In this way, they were able to get better results for unseen domains and also effectively prevent repeating or missing slots in generating the output. Further, \namecite{TranB} added a Refinement Adjustment LSTM-based component on the decoder side to control semantic information. \namecite{Dusek} re-ranked the n-best output from a sequence-to-sequence model to penalize sentences that miss the required information or add irrelevant ones. They conducted their experiments on BAGEL dataset, a short version of restaurant domain dialogue acts of \namecite{WenA} dataset. Wikibio dataset, a collection of Wikipedia Biographies and their fact tables, was introduced by \namecite{Lebret}. Then they generated the first sentence of each biography by using a neural feed-forward language model conditioned on both full structured data and structured information of the previously generated words. In addition, the authors introduced a copy mechanism for boosting the words given by the structured data. Working on the same dataset, \namecite{Liu} introduced a modified LSTM that adds a field gate into the LSTM to incorporate the structured data. Further, they used a dual attention mechanism that combines the attention of both slots names and contents of the actual slots. Also, \namecite{Sha} extended this approach and integrated a linked matrix in their model that learns the desired order of slots in the target text.

In recent years, two NLG challenges have been held. First, WebNLG, \cite{Gardent} that was based on a collection of RDF triples describing facts (entities and relations between them) from DBpedia \cite{Colin} and most of the submissions were attentional RNNs encoder-decoder based Machine Translation methods. Second, E2E \cite{DusekE2E}, that used \namecite{Dusek} as the baseline method and was based on a large dataset in the restaurant domain collected by \namecite{Novikova}. End-to-End sequence-to-sequence models were most of the submissions in that challenge \cite{Juraska,Gehrmann,Zhang,Gong}.  

\section{Approach}

The natural language generation system converts a meaning representation into an adequate and fluent description in a natural language. As shown in Figure \ref{fig: Figure1}, each meaning representation can have different forms such as a table (Figure \ref{fig: Figure1},g-i), dialogue act (Figure \ref{fig: Figure1},a-e) or RDF triples (Figure \ref{fig: Figure1}-f). In each table, there is a set of records that shows its concept, and each record has a set of field-value pairs. In dialogue acts, there are an act type and a set of slot-value pairs. RDF data type, also, is a triple consisting of Subject, Property, Object relation; that can be considered as subject and property-object pairs. In the rest of this paper, we will mention all of the meaning representation components as record and field-value pairs for consistency. 

At the test time, only records are given; without any constraint on the descriptive sentences such as the expected length or words. Thus, the goal here is to produce a system to generate a fluent sentence that accurately describes the knowledge of the given meaning representation and is also grammatically correct.  This process requires two important decisions. First, among all field-value pairs in the test records, which ones should be discussed, and conversely, which ones are either redundant or unimportant and should be ignored (content planning). Second, how to express them in a sentence (surface realization). 

The required knowledge for both decisions is learned from the training data. Training data are a set of scenarios that each scenario includes a set of records and its descriptive text.  Accordingly, a subset of fields can be selected to be expressed in the output sentence by studying the frequency of mentioning each field in all the training sentences. In this study, to generate a sentence from a given set of records, we will produce a dependency tree that includes the selected fields and a set of compatible words with them. This set of words and their dependency relations are learned from the dependency trees of the training sentences. Details of each learning step as well as sentence generation step at the test time in our proposed NLG system are elaborated in the rest of this section.

\subsection{Content Planning}

As mentioned before, each record consists of several field-value pairs; however, depending on the audience and the purpose of the NLG system, it is necessary to mention some of these pairs in the output sentence. Besides, the order of mentioning the values of these fields in the text is important. This information, that shows the main concept of a sentence, can be learned from the training data by at first aligning each training sentence with its corresponding field-value pairs. Aligning can be done by searching for these values among the words of the corresponding text. For word equal to the field value, a \textit{meaning label}, containing a pair of its equivalent record and field (in the form of [record;@field]), replaces that word. Figure \ref{fig: Figure1} shows an example of an aligned sentence for each dataset. Subsequently, the probability of mentioning each meaning label in the training sentences is calculated as follows:
\begin{equation}
\label{e1}
p(l_{i})=\frac {c(l_{i})}{N}
\end{equation}
where {\tt $N$} is the number of all meaning labels in the training data and {\tt $c(l_{i})$} is the number of times that meaning label {\tt $l_{i}$} occurs in the all aligned sentences. As a result, meaning labels with higher probabilities are more likely to be expressed in the final sentence. Furthermore, we calculate the probability of a sequence of meaning labels by making the first order Markov assumption:
\begin{equation}
\label{e2}
p(l_{0}, ..., l_{N})=p(l_{0})\prod_{i=1}^{N}p(l_{i}|l_{i-1})
\end{equation}
where {\tt $p(l_{j}|l_{i})$} is the probability that label {\tt $j$} occurs exactly right after label {\tt $i$} in a sentence and calculated as follows:
\begin{equation}
\label{e3}
p(l_{j}|l_{i})=\frac {c(l_{i},l_{j})}{\sum_{j=1}^{N}c(l_{i},l_{j})}
\end{equation}
where {\tt $c(l_{i},l_{j})$} is the number of times that label {\tt $j$} occurs exactly right after label {\tt $i$} in a sentence among all aligned training sentences. The most probable meaning labels sequence of a set of record and fields is the best order to be placed in a sentence. For testing, at first, a set of all possible meaning labels ([record;@field]) from test meaning representation, is generated. Then, using Equation (\ref{e2}), the probability of each possible sequence of meaning labels will be calculated and the most likely one is chosen as the conceptual structure of the output sentence.
\begin{figure}
    \centering
    \begin{subfigure}[b]{\textwidth}
        \includegraphics[width=\textwidth]{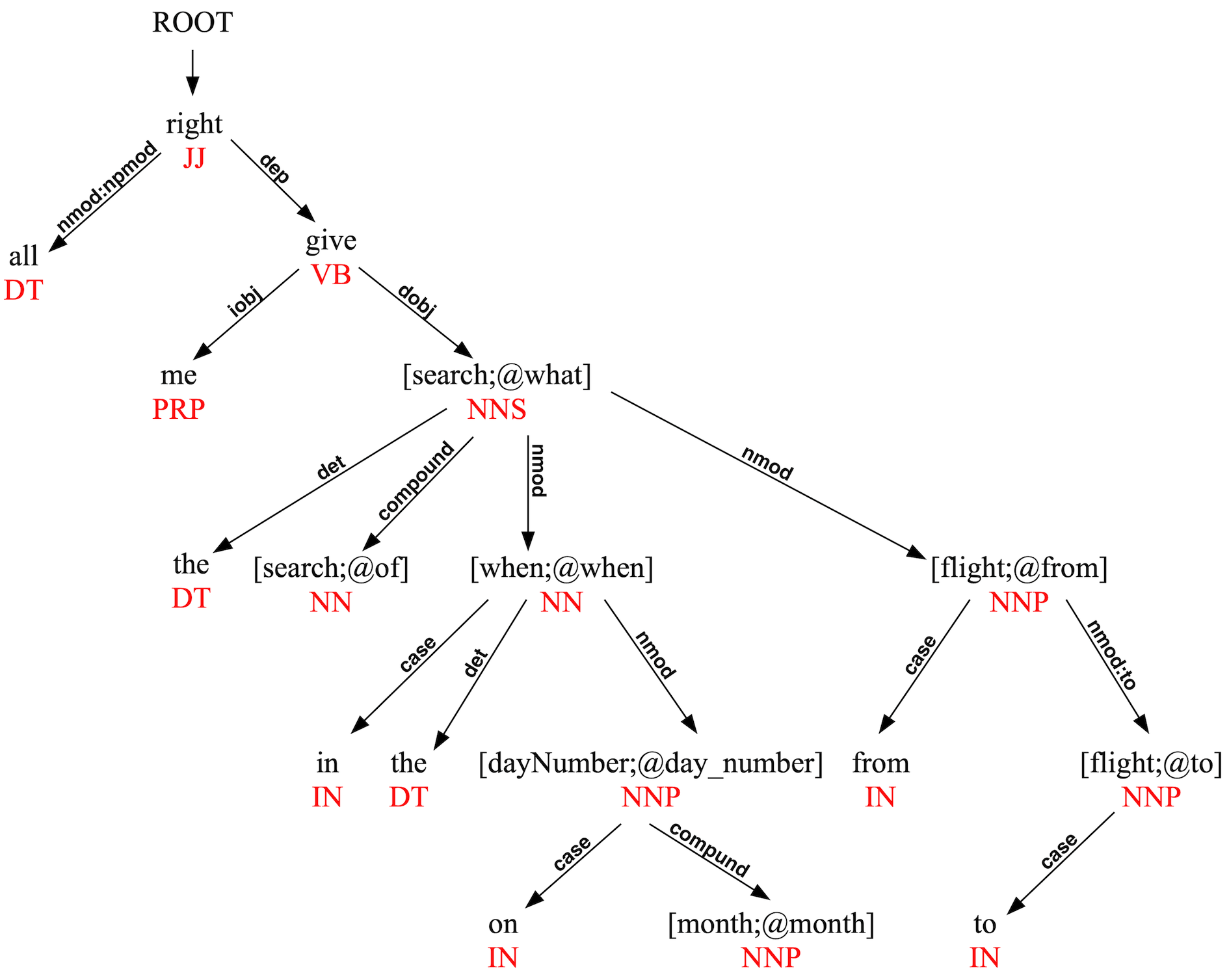} 
        \caption{Dependency tree}
        \label{fig:dp}
    \end{subfigure}
    \begin{subfigure}[b]{\textwidth}
         \includegraphics[width=\textwidth]{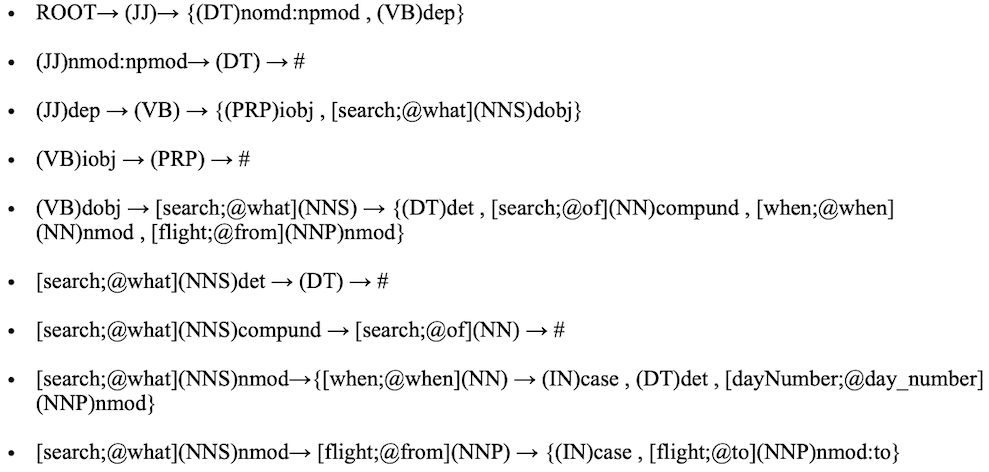}
        \caption{Extracted dependency features}
        \label{fig:dfv}
    \end{subfigure}

\captionsetup{%
    font={small}, 
    format=plain,  
    singlelinecheck=true,      
    labelsep=newline,
    labelfont=bf,      
}

\caption{ (a) An example of the dependency tree for an aligned sentence “All right give me the [search;@of] [search;@what] in the [when;@when] on [dayNumber;@day-number] [month;@month] from [flight;@from] to [flight;@to]” from the set of records shown in Figure \ref{fig: Figure1}-a. PoS tags are shown in red and edges are labeled with dependency relations.  (b) Some of the extracted dependency features of the dependency tree in (a). "\#" shows a node does not have any children.
}   
\label{fig: figure2} 
\end{figure}
 
\subsection{Dependency and Word Feature Extraction}

At the test time, only a set of records of a specific domain is given; however, just by putting its fields value together, we cannot generate a sentence and thus it is necessary to choose a set of words associated with that domain to place alongside these values. Moreover, as mentioned earlier, the descriptive sentence will be generated by producing a dependency tree consisting of the selected sequence of meaning labels and a chosen set of words. This words set and their dependency relations will be selected based on words and dependency relations in the dependency trees of the aligned training sentences. An Example of these dependency trees is shown in Figure \ref{fig: figure2}-a. As can be seen, nodes of the dependency tree are words or meaning labels with their Part of Speech (PoS) tags and the labeled edges represent their dependency relations. 
\begin{figure}
\captionsetup{%
    font={small}, 
    format=plain,  
    singlelinecheck=true,      
    labelsep=newline,
    labelfont=bf,      
}
\center
\includegraphics[width=0.4\textwidth]{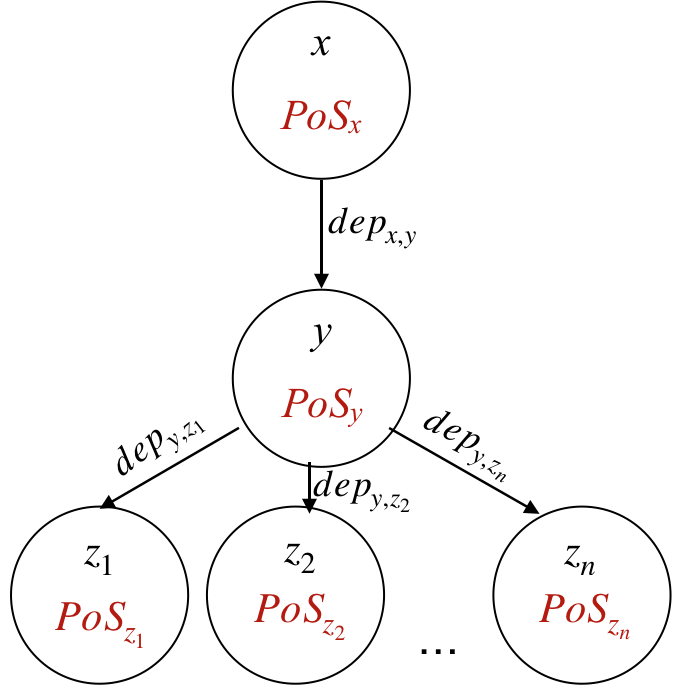} 
\caption{ Structure of the sub-tree extracted from dependency tree.  $x$ is parent of $y$  and  ${\{z_{1},...,z_{n}}\}$ are child nodes of $y$. So $x$ and $PoS_{x}$ show the meaning label of node $x$ (if it is available) and its PoS tag, and $dep_{x,y}$ shows dependency relations between nodes $x$ and $y$.}
\label{fig: Figure3}
\end{figure}

For extracting words and dependency relations from dependency trees of the training sentences at first we divide each dependency tree into a set of overlapping sub-trees. Suppose that the dependency tree contains $n$ nodes. For each node, we create a sub-tree that includes its parent and child nodes with all the dependency edges between them (Figure \ref{fig: Figure3}). It should be noted that each node in this sub-tree is a meaning label with its PoS tag. In other words, if a node was related to a word of an aligned sentence and not a meaning label, the word will be ignored and only its PoS tag will be considered.

We will represent each extracted sub-tree in a linear form as follows and refer to each of them as a "\textit{dependency feature}":
\begin{equation}
\small
L_{x},(PoS_{x}),dep_{x,y} \rightarrow L_{y},(PoS_{y}) \rightarrow {[L_{z_{1}},(PoS_{y,z_{1}}),dep_{y,z_{1}}],...,[L_{z_{n}},(PoS_{y,z_{n}}),dep_{y,z_{n}}]}
\label{eq4}
\end{equation}
where $L_{x}$ is the meaning label of node $x$ and as mentioned before, if the meaning label of a node in the dependency tree is unavailable, only the PoS tag of this node will be considered. Some examples of the extracted dependency features are shown in Figure \ref{fig: figure2}-b. Also, for each dependency feature, a probability based on its occurrence in all the dependency trees will be calculated as follows:
\begin{equation}
p(f_{i})=\frac {c(x,y,{\{z_{1},...,z_{n}}\})}{\sum_{w} c(x,y,{\{w}\})}
\label{eq5}
\end{equation}
where $c(x,y,{\{z_{1},...,z_{n}}\})$ is the number of times that $x$ and ${\{z_{1},...,z_{n}}\}$ are the parent and child nodes of node $y$ and the denominator of the fraction is the total number of times that $x$ is the parent node of $y$ in the set of the extracted dependency features from all dependency trees. Suppose that the training data contains $m$ sentences with an average $n$ words. Then for each dependency tree of an aligned sentence, $n$ sub-trees and consequently $n$ dependency features will be extracted. Therefore, the total number of dependency features extracted from all training data in the general case has the order of $O(m \times n)$\footnote{It should be noted that due to a large number of repetitive phrases and words in the training sentences, many of the extracted features are repetitive. So the actual number of features will be much lower.}. These dependency features are like blocks of information that will be concatenated together at the test time to create a new dependency tree.

It should be noted that a word can have various number of arguments (adjectives, adverbs, prepositions, etc.) based on its syntactic role in a sentence. Even a single headword with the same syntactic role can have different numbers of the dependent word in different sentences based on the other constitutive words of the sentences and their concepts. Therefore, the number of arguments for each word should be taken into account at the time of adding new words and dependency relations to the dependency tree of the test data; because over- or under- assigning dependents to a headword in the dependency tree generates a grammatically incorrect sentence. Accordingly, we have chosen this format for dependency features to also implicitly learn the appropriate number of dependent words for each word based on its PoS tag and dependency relation with its parent node. In addition, using Equation (\ref{eq5}), the most likely number of dependent words for each syntactic role is also determined. Therefore, in the process of adding a word by concatenating its corresponding sub-tree to the dependency tree of the test data, using this information results in an appropriate number of arguments for it. 

As mentioned before, for generating a sentence we also need a lexicon containing all words in the training sentences. Hence, a set of "\textit{words feature}" similar to the feature (\ref{eq4}) for words in each sub-tree is created; with the difference that the word of each non-meaning label node in the sub-tree is also considered in addition to its PoS tag. Moreover, a 3-gram language model is also generated according to the words and their PoS tags in the training sentences.

\begin{algorithm}
\caption{The pseudo-code for producing dependency tree algorithm.}
\label{AL1}
\begin{algorithmic}[1]
 \STATE \textbf{Initialization:} Set $T_{dep} \leftarrow \{\}$ , $Candidate \leftarrow \{\}$ , $Score(T_{dep})=0$
 \STATE \textbf{Input:} The extracted dependency feature set $F$ and an ordered sequence of meaning labels $L={\{l_{1},...,l_{n}\}}$
 \STATE \textbf{Output:} Dependency Tree $T_{dep}$ , $Score(T_{dep})$
 \STATE \textbf{Begin}
 \FORALL{ $f(p_{1} \rightarrow p_{2} \rightarrow p_{3}) \in F$}
    \IF{$p_{1}=$ROOT and $l_{1} \in( p_{2}$ or $p{3})$ }
        \STATE Check if all the other meaning labels  in the $f$ are in $L$ and follow the specified order then add $f$ to $Candidate$ set
    \ENDIF
    \ENDFOR
 \STATE $T_{dep} \leftarrow \argmax_{\theta \in Candidate} p(\theta)$
 \STATE $Score(T_{dep})=Score(T_{dep})+\log(p(\theta))$
 \STATE remove all meaning labels in $T_{dep}$ from $L$
\WHILE{all child node $t$ in $T_{dep}$ are not visited}
        \FORALL{ $f(p_{1} \rightarrow p_{2} \rightarrow p_{3}) \in F$}
         \IF{$(l_{parent_{t}},PoS_{parent_{t}},dep_{parent_{t},t})=p_{1}$ and $(l_{t},PoS_{t})=p_{2}$}
                \STATE Check if all the other meaning labels in the $f$ are in $L$ and follow the specified order then add $f$ to $Candidate$ set
          \ENDIF
         \ENDFOR           
  \STATE $T_{dep} \leftarrow \argmax_{\theta \in Candidate} (p(\theta),count(labels \in \theta ))$
   \STATE  $t \leftarrow visited$
   \STATE $Score(T_{dep})=Score(T_{dep})+\log(p(\theta))$
   \STATE remove all meaning labels in $\theta$ from $L$
\ENDWHILE   
\end{algorithmic}
\end{algorithm}
\subsection{Dependency Tree Producing}

At this point, for a given set of records as test data, the main structure of its corresponding descriptive sentence will be created. This main structure of the output sentence is represented by the dependency relations between its words. Therefore, after selecting an ordered sequence of meaning labels from test data in the content planning step, a dependency tree consisting of these labels and a set of additional words is created. This dependency tree, as shown in Algorithm \ref{AL1}, is produced incrementally and from top to bottom by using the set of dependency features extracted from training data. Given the dependency feature set $F$ and an ordered meaning labels sequence $L$, this algorithm starts by finding those dependency features that contain the ROOT node and the first selected meaning label (lines 5-9). Among all the features in $F$ that matched with these conditions, the most probable one is chosen and its sub-tree is added to an empty tree (line 10). Moreover, the score of the dependency tree is updated and the meaning labels in the added sub-tree also are removed from $L$ (lines 11-12). 

\begin{figure}
    \begin{subfigure}[b]{0.5\textwidth}
        \includegraphics[width=0.9\textwidth]{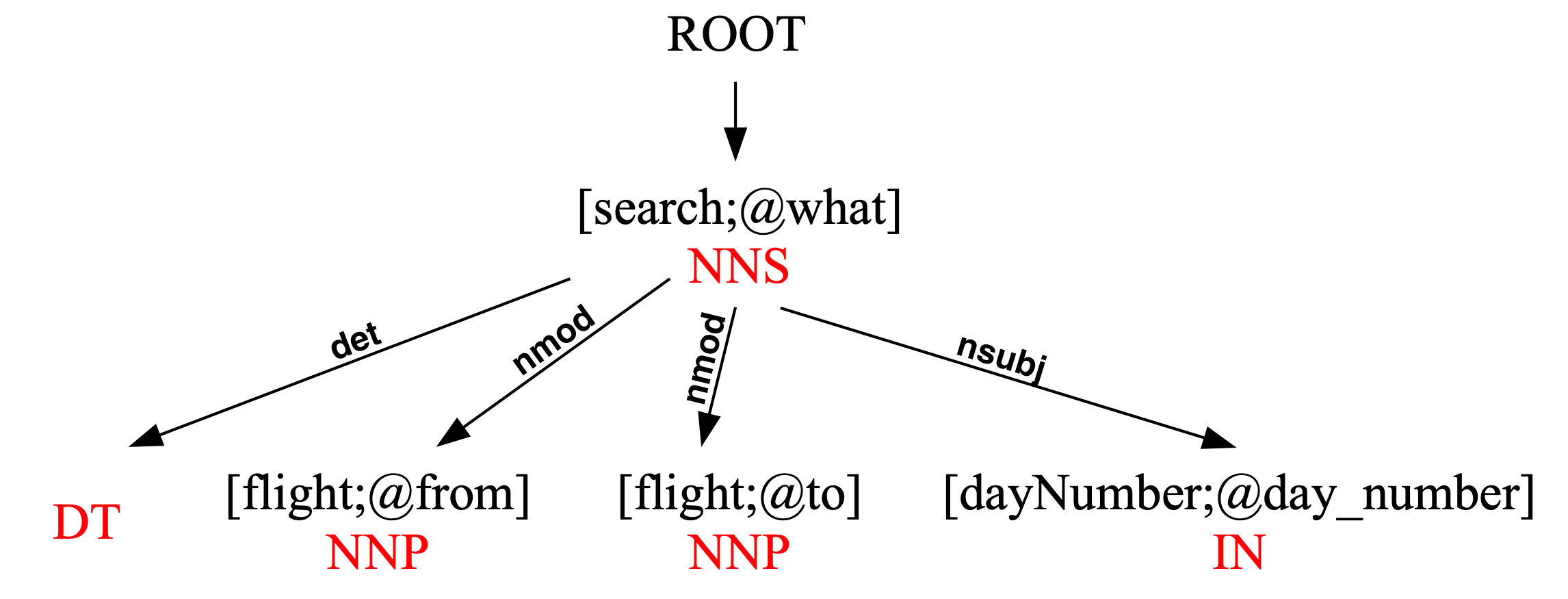} 
        \caption{}
    \end{subfigure}
        \begin{subfigure}[b]{0.5\textwidth}
        \includegraphics[width=0.9\textwidth]{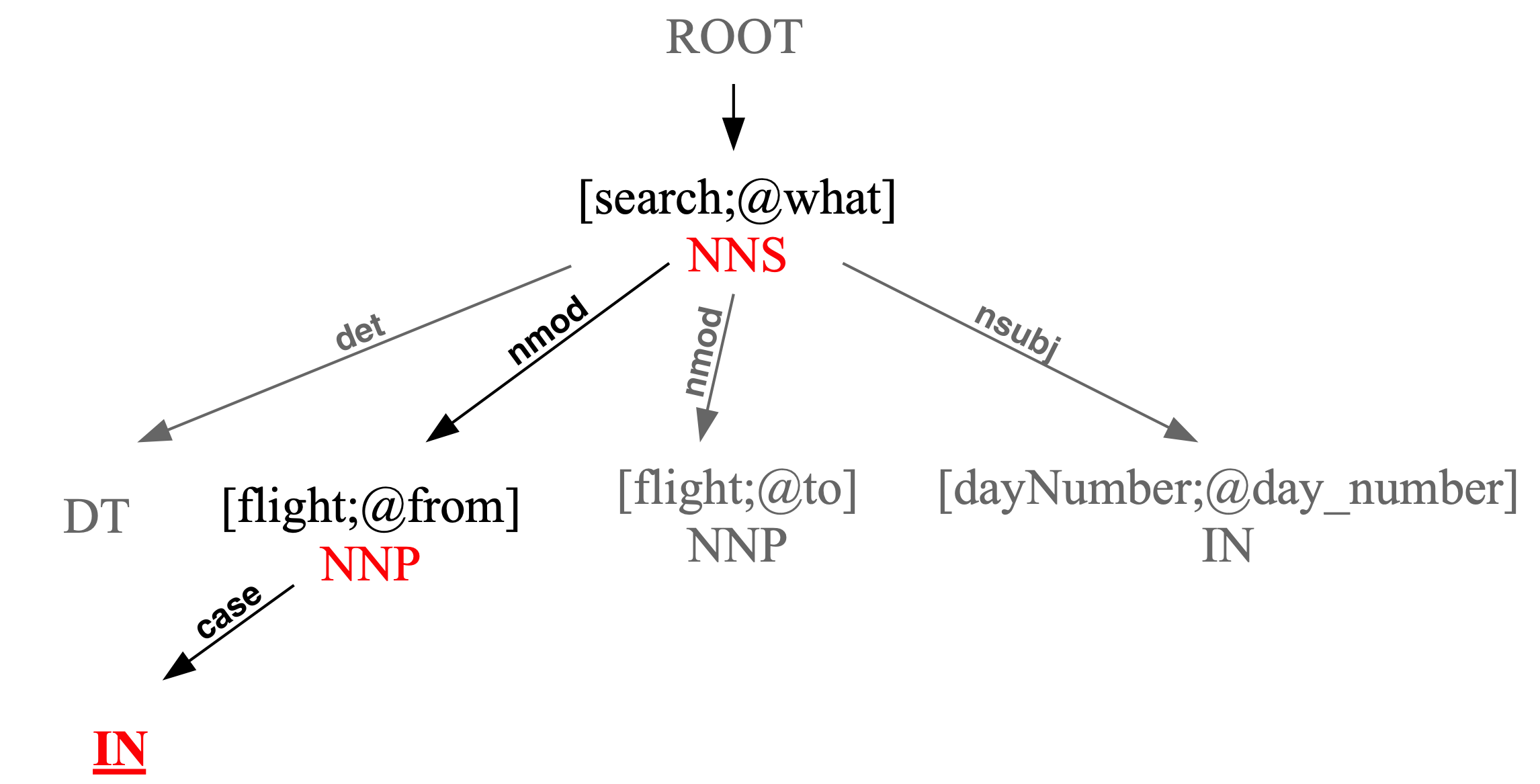} 
         \caption{}
    \end{subfigure}
        \begin{subfigure}[b]{0.5\textwidth}
        \includegraphics[width=0.9\textwidth]{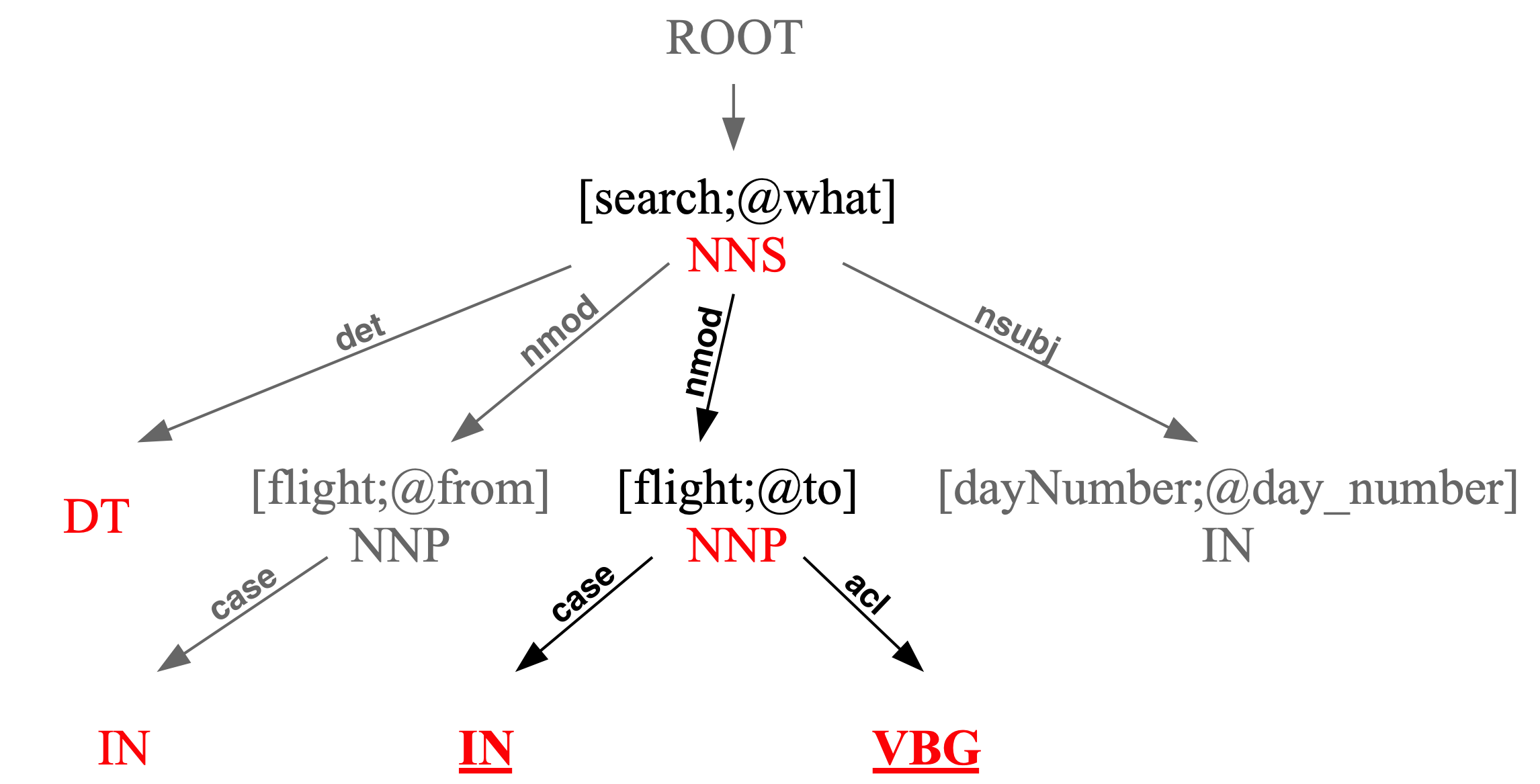} 
                 \caption{}
    \end{subfigure}
        \begin{subfigure}[b]{0.5\textwidth}
        \includegraphics[width=0.9\textwidth]{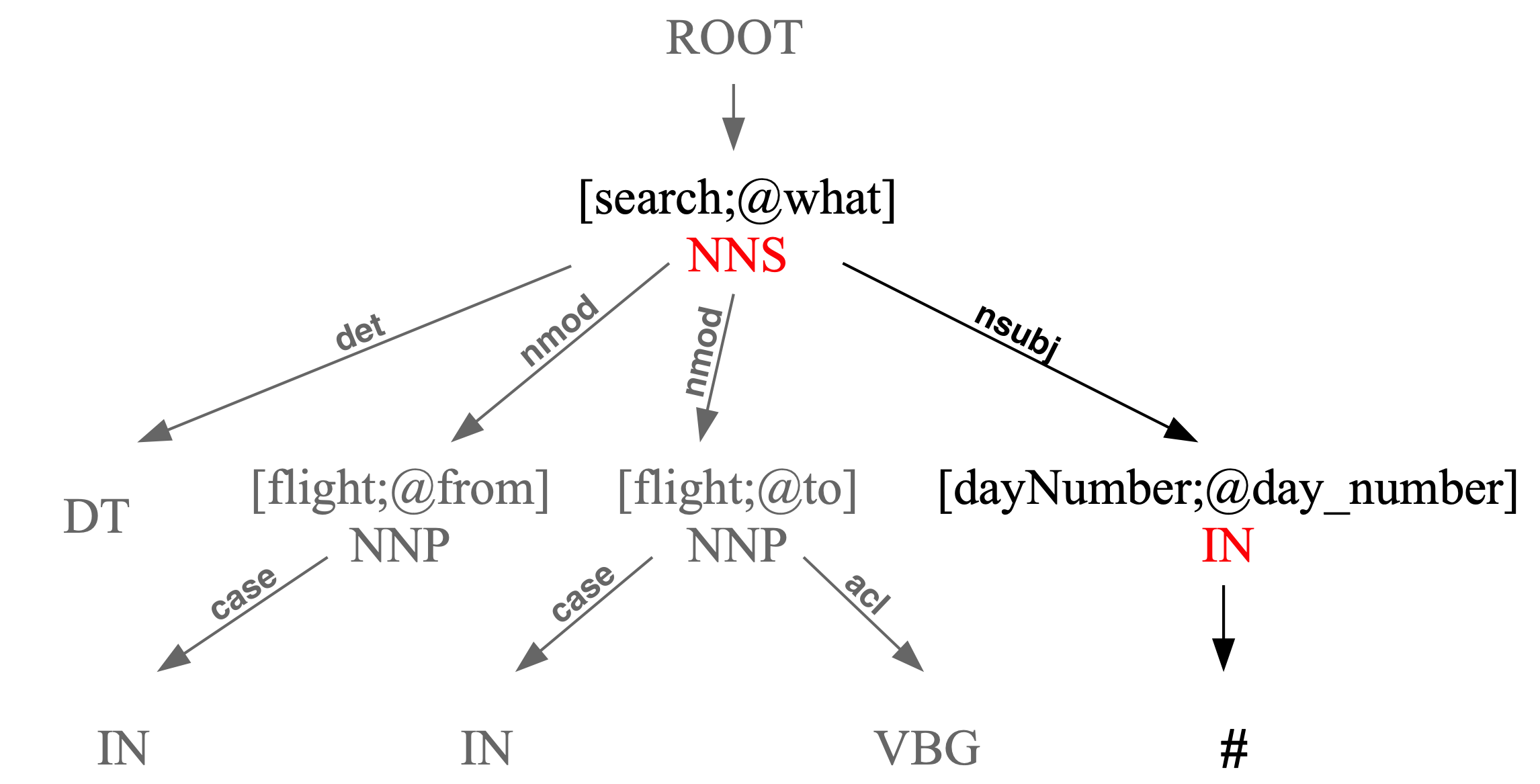} 
                 \caption{}
    \end{subfigure}
        \begin{subfigure}[b]{0.5\textwidth}
        \includegraphics[width=0.9\textwidth]{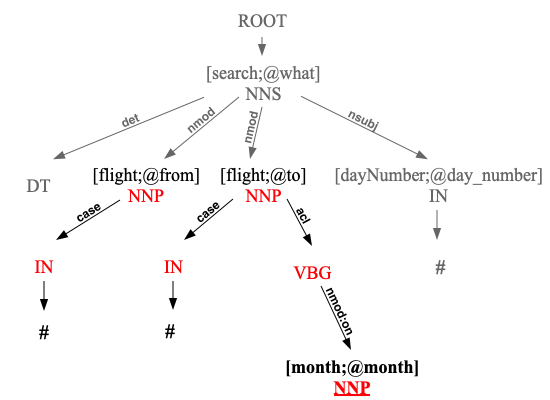} 
                 \caption{}
    \end{subfigure}
        \begin{subfigure}[b]{0.5\textwidth}
        \includegraphics[width=0.9\textwidth]{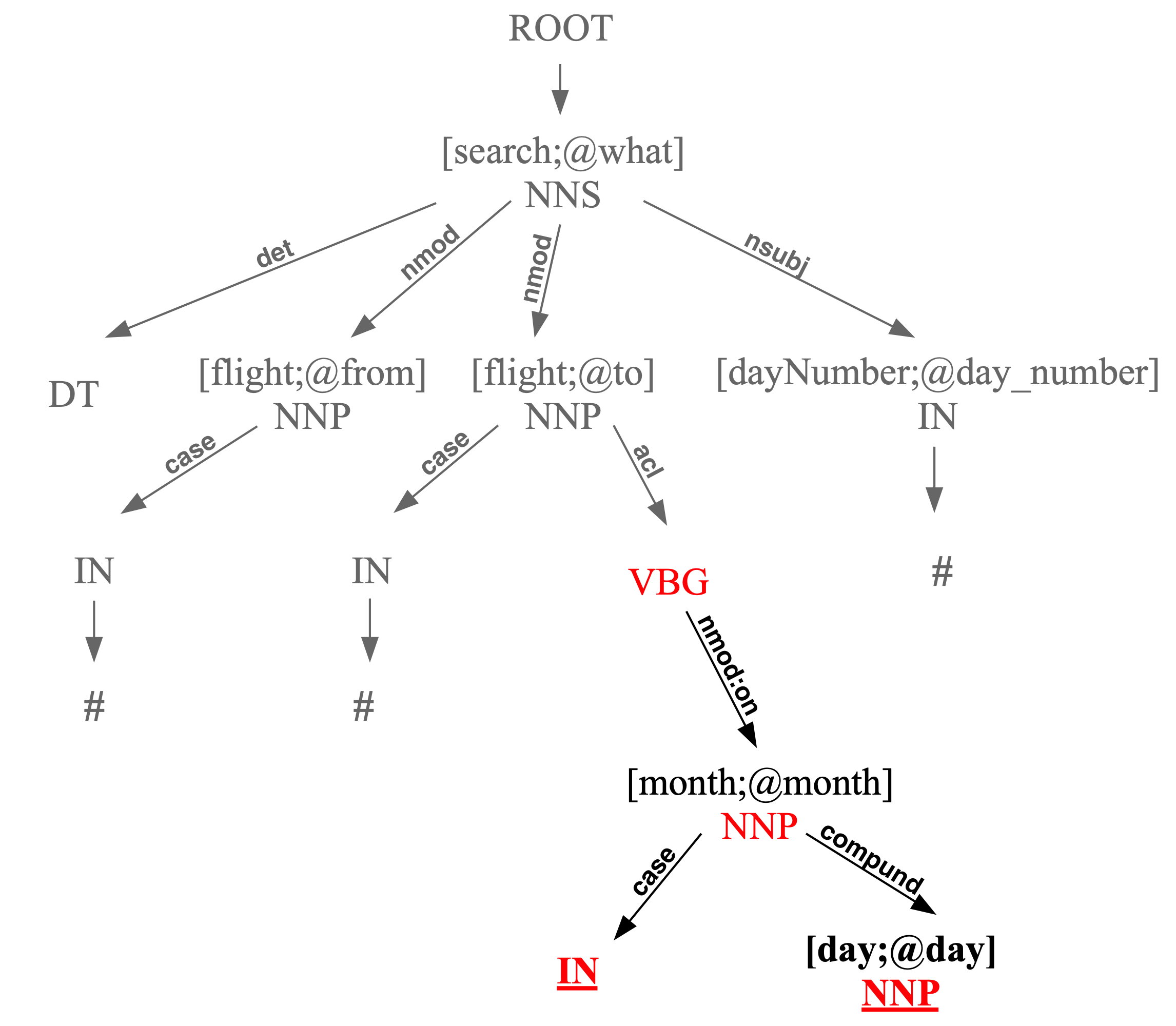} 
                 \caption{}
    \end{subfigure}
            \begin{subfigure}[b]{0.5\textwidth}
        \includegraphics[width=0.9\textwidth]{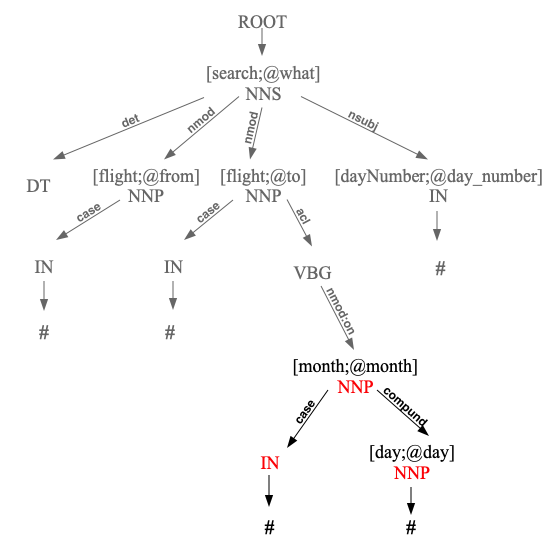} 
                 \caption{}
    \end{subfigure}
            \begin{subfigure}[b]{0.5\textwidth}
        \includegraphics[width=0.9\textwidth]{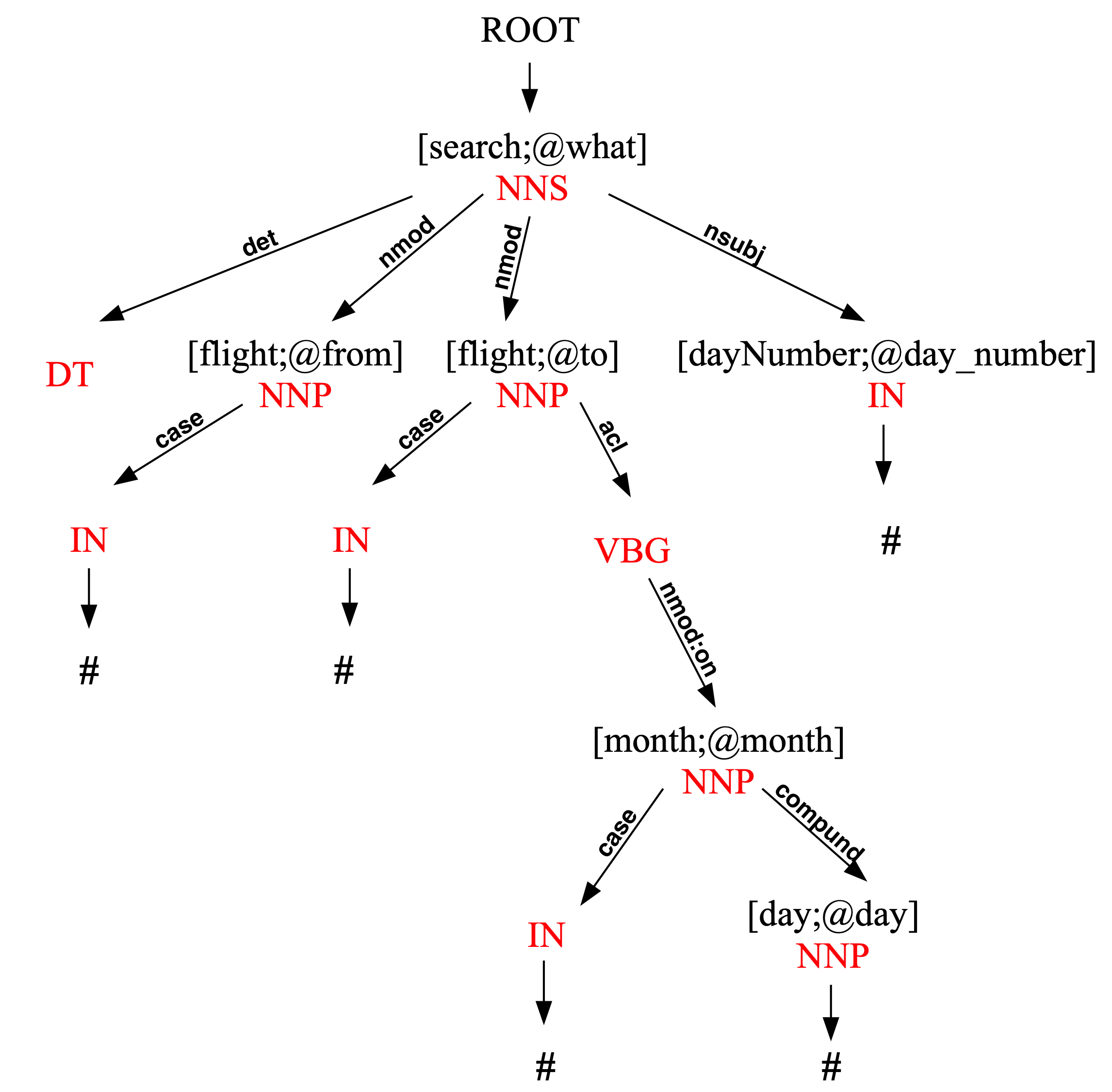} 
                 \caption{}
    \end{subfigure}
    \captionsetup{%
    font={small}, 
    format=plain,  
    singlelinecheck=true,      
    labelsep=newline,
    labelfont=bf,      
}
\caption{Producing dependency tree for an ordered sequence of meaning labels {[search;@what], [flight;@from], [flight;@to], [day;@day], [month;@mont], [dayNumber;@day-number]}. "\#" means there is not any child node and bold underlined nodes are new child nodes that are added to the tree.}
\label{fig: Figure5}
\end{figure}

Thereafter, for each child node $t$ in all next levels of the produced tree and from left to right, the PoS tags, meaning labels and dependency relations of node $t$ and its parent node are compared with the first two parts of all features in the extracted dependency feature set and consequently a set of candidate dependency features is created (lines 13-18). Among all candidate dependency features, a dependency feature that not only has a high probability, but also the largest number of meaning labels (in the determined order), is chosen and all nodes in the last level of it are added to the produced dependency tree as child nodes of $t$ and their dependency relations are added as the labeled edges between $t$ and new child nodes (line 19). Eventually, the score of the produced dependency tree, $Score(T_{dep})$, is updated as follows:
\begin{equation}
Score(T_{dep})=Score(T_{dep})+log(p(\theta))
\label{eq6}
\end{equation}
where $p(\theta)$ is the probability of the added dependency feature $\theta$ to the produced dependency tree. Furthermore, all meaning labels in the dependency feature $\theta$ are removed from the meaning labels sequence L (lines 21,23). This process will continue until all child nodes are visited and thus no more nodes can be added to the dependency tree. In the end, the produced tree can be considered as the dependency tree of the descriptive sentence for the test data. For generating more than one sentence for each test data, we can use beam search, which means at each step the $B$ most probable features can be chosen; therefore, in the end, at least $B$ different dependency trees (because of choosing $B$ ROOT features) and thus at least $B$ different sentences with different number of nodes can be generated. Then, the trees with higher scores will be chosen as the final dependency trees for the test data. Suppose the size of the extracted dependency feature set is $n$. Since each dependency feature has three levels and comparison is done for all levels, the complexity of the dependency tree producing algorithm in the worst case has the order of $O(Bn^{3})$. Figure \ref{fig: Figure5} illustrates the process of producing a sample dependency tree.

Using three-level sub-trees and dependency relationships for generating a sentence has four reasons: first, as mentioned earlier, adding the correct number of dependents to a headword is one of the main challenges in producing the dependency tree. Choosing the most likely sub-tree with adequate number of dependents and adding its dependent nodes to the produced dependency tree can prevent the problem of an insufficient number of arguments. Second, the correctness of the tree can be assured, because each added node was selected with respect to the dependency relations and the PoS tags of its parent and ancestor nodes. More precisely, we generate a dependency tree based on not the words themselves, but on their PoS tags, their meaning labels (if it is available) and the dependency relationships between them. Hence, there may be many nodes with the same PoS tags and no meaning labels. What sets these nodes apart is the history of their dependency relations with their parent nodes. It would be difficult to identify the correct node for adding the child nodes only based on its PoS tag and without considering its history. But using the dependency history, embedded in the three-level structure, selecting the appropriate node as a place to add new nodes can be done more accurately. Third, by generating a sentence using the dependency tree, there is no need to do calculations to predict the length of the resulting sentence because the number of words in the resulting sentence is equal to the number of nodes in its corresponding dependency tree. Fourth, a concept can be expressed in different sentences with different sets of words. So for the same concept, there will be several different dependency trees with the same meaning labels but a different set of words. By concatenating sub-trees that are extracted from these trees, a new dependency tree with a combination of their words and dependency relationships can be produced. This new dependency tree will lead to generating a new sentence that sometimes is different from all the sentences in the training data. 

\subsection{Surface Realization}

Once the dependency tree has been produced, all that remains is lexicalizing nodes of the tree and then obtaining an ordering of words based on it. As shown before, each dependency tree has two types of nodes; the meaning label nodes and the PoS tag nodes. Accordingly, two methods are used to replace dependency tree nodes with their corresponding words. For assigning words to the PoS tag node $x$, the PoS tags, meaning labels (if it is available) and dependency relations of node $x$ and its parent node, as well as child nodes, are compared with all extracted word features of the training data. In the case of matching, words of the most probable feature are assigned to their corresponding nodes in the dependency tree. Since more than one word feature can match each sub-tree of the produced dependency tree, it is possible to use either word with the highest probability or make different copies of the produced dependency tree and use words of different possible features. In the latter case, more than one corresponding sentence can be generated from the produced dependency tree. Moreover, the meaning label nodes are replaced by their values in the test data. In this way, a complete dependency tree is created for the descriptive sentence of test data. 

\begin{figure}
\captionsetup{%
    font={small}, 
    singlelinecheck=true, 
    format=plain,             
    labelsep={colon},         
}
\center
\includegraphics[width=0.7\textwidth]{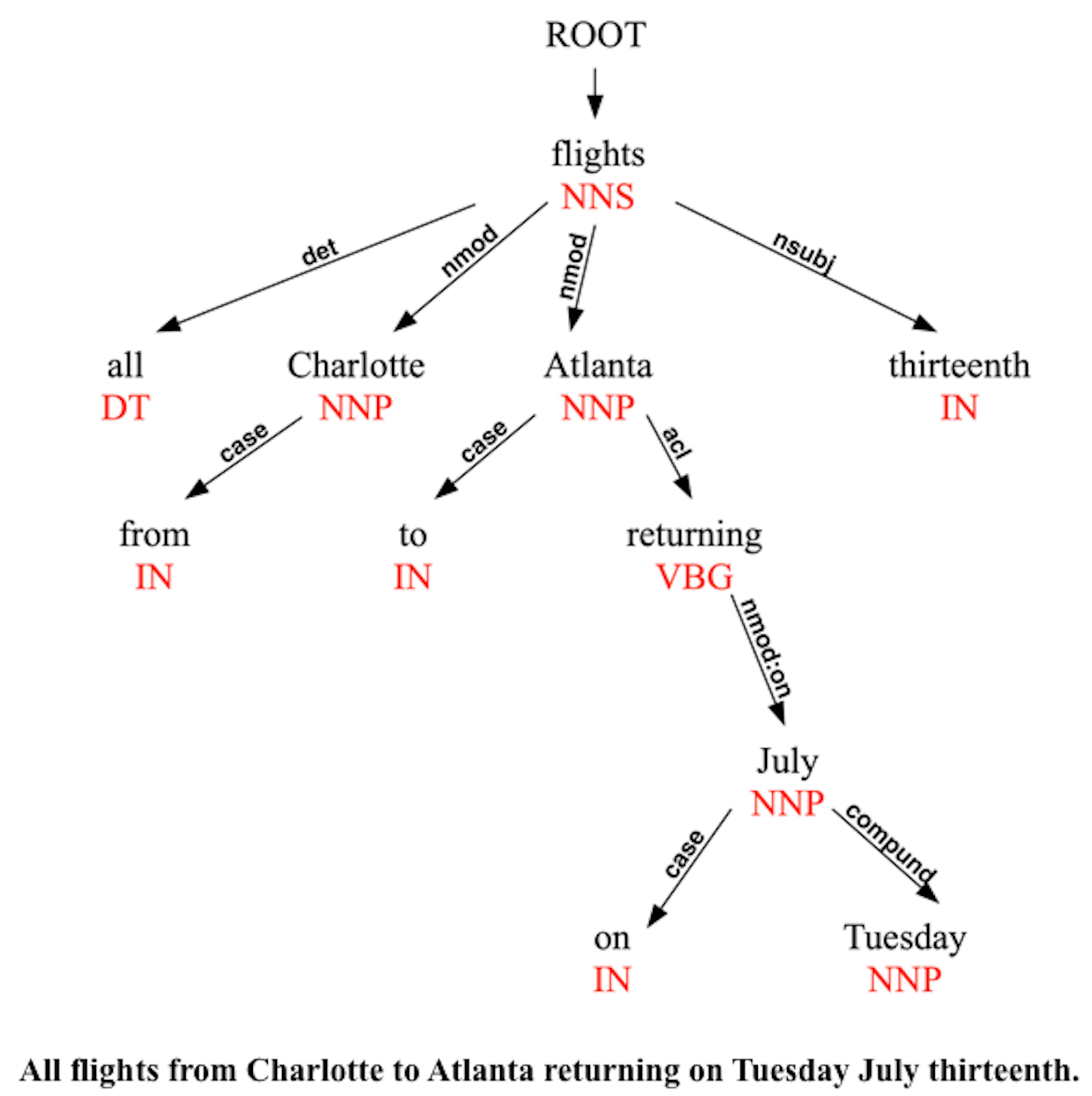} 
\caption{The final dependency tree and its corresponding sentence for the test records set {search(what=flight), flight(from=Charlotte, to=Atlanta), dayNumber(day-number=thirteen), month(month=July), day(day=Tuesday)}.}
\label{fig: Figure6}
\end{figure}

The final step is to extract a sequence of words from this dependency tree. For this purpose, the completed dependency tree is traversed in a depth-first in-order manner.  Then, for each sub-tree, permutations of a headword and its dependents are generated and based on the 3-gram learned language model from the training data (including words and their PoS tags), the most likely permutations are chosen. These permutations are concatenated and a phrase is created. The generated phrase will be replaced by its corresponding sub-tree. For example, in the sub-tree consisting of \textit{"July(NNP), On(IN), Tuesday(NNP)"} in Figure \ref{fig: Figure6}, the most likely sequence is \textit{"On Tuesday July"}, so this phrase will be replaced by this sub-tree and as a child node of \textit{"returning(VBG)"} with the PoS tag of \textit{"July"}. This process will be continued with generating permutations of the next headword, its dependents and the first and last words of any child phrase until all nodes of the dependency tree participate in the final sequence of words. In the end, all PoS tags will be removed. The final score of each sentence is calculated as the logarithmic sum of the probability of the used permutations; these scores will be used to select the best output sentences. Examples of the completed dependency tree and its corresponding sentence for a test scenario are shown in Figure \ref{fig: Figure6}.

\section{Experiments}

\subsection{Datasets}

We conducted the experiments on nine different NLG benchmark datasets: RoboCup \cite{Chen}, Atis \cite{KonstasA}, and WEATHERGOV \cite{Liang} in table format, Restaurant, Hotel, TV and Laptop \cite{WenA,WenD} in dialogue act format and also datasets of E2E\citep{Novikova} and WebNLG\citep{Colin} challaenge. The RoboCup dataset consists of 1,539 scenarios of four RoboCup final games in 2001-2004. This is the simplest dataset available for this task, since it has a small vocabulary and short sentences with simple syntax (e.g., a transitive verb with its subject and object). Given the relatively small size of this dataset, we follow the evaluation methodology of the \namecite{Chen} and \namecite{Angeli} by performing four-fold cross-validation,  whereby we train on three games (approximately 1000 scenarios) and test on the fourth. The WEATHERGOV dataset consists of 29,528 weather scenarios for 3,753 major US cities (collected over four days). Following \namecite{Angeli} and \namecite{KonstasA}, we use WEATHERGOV training, development, and test splits of size 25000, 1000, and 3528, respectively. The ATIS dataset consists of 5,426 scenarios. These are transcriptions of spontaneous utterances of users interacting with a hypothetical online flight booking system. Following \namecite{KonstasA}, the proposed model is trained on 4,962 scenarios and tested on ATIS NOV93 which contains 448 examples. 

The Restaurant, Hotel, TV and Laptop datasets consist of 5373, 5192, 7035 and 13242 dialogue acts and their corresponding utterances for each domain, respectively; all used in the ratio 3:1:1 of the train, validation, and test set following \namecite{WenA,WenC,WenB,WenD} evaluation methods. The E2E dataset consists of 51426 references for 6039 distinct MRs in the restaurant domain, which is 10 times bigger than previous datasets. For using the E2E dataset, we split the data into training, development and test sets in a roughly 82:9:9 ratio following \namecite{DusekE2E}. The WebNLG \cite{Colin} dataset consists of 25,298 data and text pairs from 15 different domains. The data units are sets of RDF triples extracted from DBPedia and the texts are sequences of one or more sentences verbalizing these data units. We split this data into training, development and test sets in a roughly 89:5:6 ratio following \namecite{Gardent}. Statistics of all datasets are shown in Table \ref{table1}.

\begin{table}
\begin{center}
\captionsetup{%
    font={small}, 
    format=plain,  
    singlelinecheck=true,      
    labelsep=newline,
    labelfont=bf,      
}
\caption{Datasets Statistics. {\tt Sent/Scenario} shows an average number of sentences in each scenario, {\tt |Sent|} shows an average length of each sentence, {\tt Attr.} shows the total number of record type in each dataset and {\tt |Attr.|/Scenario} shows an average number of record types in each scenario.}
\label{table1}
\fontsize{8.5}{15}\selectfont
\begin{tabular}{>{\bfseries} +m{2cm} >{\centering\arraybackslash}^m{1.2cm} >{\centering\arraybackslash}^m{1.2cm} >{\centering\arraybackslash}^m{0.7cm} >{\centering\arraybackslash}^m{0.7cm} >{\centering\arraybackslash}^m{1.2cm}>{\centering\arraybackslash}^m{0.7cm} >{\centering\arraybackslash}^m{0.7cm} ^c}
\hline
\rowstyle{\bfseries}
 Dataset &  Scenario & Sent/ \newline Scenario  &  |Sent| &  Attr. &  |Attr.|/ \newline Scenario &  Train & Test &  Validation \\ \hline
RoboCup & 1539 & 1 & 5.7 & 9 & 2.4 & \multicolumn{3}{>{\centering\arraybackslash} c} {4-fold Cross Validation} \\
Atis & 5962 & 1 & 11.2 & 19 & 3.79 & 4962 & 448 & - \\
WEATHERGOV & 29528 & 3.25 & 9.29 & 36 & 5.8 & 25000 & 3528 & 1000 \\
Restaurant & 5192 & 1 & 8.82 & 22 & 2.86 & 3114 & 1039 & 1039 \\
Hotel & 5373 & 1 & 9.12 & 22 & 2.66 & 3223 & 1075 & 1075\\ 
TV & 7035 & 1 & 21.44 & 29 & 4.78 & 4221 & 1407 & 1407\\ 
Laptop & 13242 & 1 & 23.42 & 33 & 4.86 & 7944 & 2649 & 2649\\ 
E2E & 51426 & 1.42 & 20.1 & 8 & 5.37 & 42061 & 4672 & 4693\\ 
WebNLG & 25298 & 3.5 & 22.69 & 373 & 4.86 & 22592 & 1187 & 1519\\ \hline
\end{tabular}
\end{center}
\end{table}
\subsection{Experiments on tabular format datasets}
\subsubsection{Experimental Setup}
We used CoreNLP\footnote{http://nlp.stanford.edu:8080/corenlp} dependency parser provided by Stanford University for extracting dependency tree from the aligned training sentences. For generating dependency trees for each test scenario, we used Beam search width $B={\{1,5,10,15,20}\}$ and after generating sentences from the trees, for each $B$, we selected the most probable sentence as the output. The corpus-based-BLEU-4 metric was used for the objective evaluation by using BLEU implementation in the Natural Language Toolkit (NLTK)\footnote{https://www.nltk.org/} \cite{Loper}. In addition, for comparing WEATHERGOV results with previous works, we used BLEUc that is modified BLEU score and does not penalize numerical deviations of at most five. As gold-standard references, the utterances in the RoboCup, Atis and WEATHERGOV test scenario were used. We compared our proposed models against with strong baselines including four statistic models \citep{Kim,Angeli,KonstasA,KonstasB}, RNN model with encoder-aligner-decoder architecture \cite{Mei} and a Neural Machine Translator system (NMT)\cite{LeeC} that are trained on Atis, WEATHERGOV and RoboCup datasets.

To evaluate the generated sentences by our proposed system in terms of informativeness, naturalness, and quality, we also ran Human Evaluation. To do this, we selected randomly 20 test scenarios for each dataset and used gold-standard references as baselines. Also, as judges, we used 10 students from the University of Amsterdam, whose native language was English (Fleiss’s $\kappa$ =0.69, Krippendorff’s $\alpha$ =0.72). Each judge was shown all randomly selected test scenarios and their baselines, turn by turn. At each turn, a sentence was presented to them, without specifying whether this sentence is the output of our system or the baseline, and the judge was asked to score each sentence in terms of each measuring factor separately of 1 to 5. Here, informativeness is defined as whether the generated sentence provides all the useful information from the test scenario, naturalness is defined as whether the utterance could have been produced by a native speaker and quality is defined as how they judge the overall quality of the generated sentence in terms of its grammatical correctness, fluency, and adequacy.

\subsubsection{Individual domain experiment}
In this experiment, we trained our proposed NLG system on each dataset separately and then compared it against the baseline systems. The objective evaluation results for different $B$ values are shown in Table \ref{table2} and overall comparisons are shown in Table \ref{table3}. As can be seen, in $B=1$ which is actually a greedy search, the BLEU scores are low and by using larger values for $B$ the scores are also increased. This is contrary to the results of \namecite{Mei,Angeli} about the superiority of greedy search over the beam search. Because in our proposed method, to achieve greater diversity in the output sentences, we construct the structure of the sentences from the basic information, including PoS tags, meaning labels and dependency relationships between them and in the next steps we add the words. When constructing dependency trees, if we act greedily and consider only one sub-tree with the most likely dependency feature with the maximum number of meaning labels at each step, the same tree will be generated for all the same meaning representations. Also, in realization step, this is possible that no proper words features for a sub-tree are found and eventually no sentences are produced. But by using beam search and based on $B$ values, more trees are generated and also in the realization step, more choices are given and eventually different sentences will be produced.

As it is shown in Table \ref{table3} by using dependency relations instead of context-free grammars (CFG) \cite{Kim}, probabilistic context-free grammars (PCFG) \cite{KonstasA,KonstasB} or making decisions based on n-grams only and then using log-linear of the results \cite{Angeli} for generating sentences, our system outperforms all the state-of-the-art systems in term of the BLEU. Because in our proposed system, words are considered as individual components; therefore, based on the diversity of dependency relationships they can have with other types of words, they can be placed in the different parts of a sentence by using the variety of the PoS tags; Unlike in CFG and PCFG that words are considered as part of a phrase and have fixed PoS tags. Also, our system achieves a comparable result with the baseline system that used RNN \cite{Mei,LeeC}. 

Our BLEU score has a large jump compared to previous works. Because of the special properties of the RoboCup dataset like having short sentences (average 5.7 words) and thus simple grammatical sentences, the small vocabulary set (only 244 words), only one meaning label in each sentence and having almost a few fixed templates for each meaning label's sentences, it has a limited set of dependency relation types. Hence, our proposed approach, as we expected, was not challenged by learning the dependency structure of sentences and generating the output sentences. We do not consider this as the strength of our system, but rather the simplicity of the dataset. Examples of the generated sentences for a test scenario from RoboCup dataset are shown in Table \ref{table4}.  

The mean scores of each evaluation factor results from Human Evaluation are shown in Table \ref{table5}. Since in our proposed system all meaning labels are forced to exist in the produced dependency tree, the mean score for informativeness factor is 5, except for RoboCup dataset; because in this dataset, each sentence corresponds to one meaning label and 60\% of the training sentences are corresponding to PASS record type. So, as a result, at the test time, since test scenarios consist of multiple record types including PASS, a few of the sentences, that have less training sample compare to the other record type (like Offside or BallStopped that have less than 2\% training sample in total), are wrongly generated for this record. Furthermore, as can be seen in Table \ref{table5}, the sentences generated by our system received high scores from the judges for the naturalness and quality factors. 
\begin{table}
\begin{center}
\captionsetup{%
    font={small}, 
    format=plain,  
    singlelinecheck=true,      
    labelsep=newline,
    labelfont=bf,      
}
\caption{Performance of the proposed system on three tabular format datasets based on different Beam widths (B) and number of candidate sentences (C) in term of the BLEU score.}
\label{table2}
\fontsize{8.5}{15}\selectfont
\begin{tabular}{+l ^c ^c ^c^c^c}
\hline
\rowstyle{\bfseries}
dataset & B=1, C=1 & B=5, C=5 & B=10, C=10 & B=15, C=15 & B=20, C=20\\ 
 \hline
RoboCup & 74.04 & 81.93 & 86.05 & 89.03 & 90.15\\
Atis & 38.91 & 46.72 & 51.09 & 54.21 & 55.17\\
WEATHERGOV & 49.87 & 55.11 & 59.62 & 62.01 & 63.17\\
 \hline
\end{tabular}
\end{center}
\end{table}
\begin{table}
\begin{center}
\captionsetup{%
    font={small}, 
    format=plain,  
    singlelinecheck=true,      
    labelsep=newline,
    labelfont=bf,      
}
\caption{Performance of different models on three tabular format datasets in term of the BLEU score; bold denotes the best model\textsuperscript{*}.}
\label{table3}
\fontsize{8.5}{15}\selectfont
\begin{tabular}{+l ^c ^c ^c^c}
\hline
\rowstyle{\bfseries}
\multirow{2}[3]{*}{Method} &   \multicolumn{1}{c}{\textbf{RoboCup}} &
\multicolumn{1}{c}{\textbf{Atis}} & \multicolumn{2}{c}{\textbf{WEATHERGOV}}  \\ \cmidrule(l){2-5} 
 & \multicolumn{1}{c}{\textbf{BLEU}}
 & \multicolumn{1}{c}{\textbf{BLEU}}
 & \multicolumn{1}{c}{\textbf{BLEU}}
 & \multicolumn{1}{c}{\textbf{BLEUc}} \\
 \hline
 CFG\cite{Kim} & 47.27 & - & - & -\\
 log-linear\cite{Angeli} & 38.04 & 26.77 & 38.40 & 51.50\\
 PCFG\cite{KonstasA} & 30.90 & 30.37 & 33.70 & -\\
 PCFG\cite{KonstasB} & - & - & 36.54  & -\\ 
 RNN\cite{Mei} & - & - & 61.01 & 70.39\\
 NMT\cite{LeeC} &  36.62 & - & 36.93 & \textbf{78.90}\\
 \hline
 Our System & \textbf{90.15} & \textbf{55.17} & \textbf{63.17} & 75.49 \\
 \hline
 \multicolumn{4}{p{0.8\textwidth}}{\textsuperscript{*}\small \textit{Notes. The BLEU scores for baseline systems are reported scores by their authors.}}
\end{tabular}
\end{center}
\end{table}
\begin{table}
\begin{center}
\captionsetup{%
    font={small}, 
    format=plain,  
    singlelinecheck=false,      
    labelsep=newline,
    labelfont=bf, 
    justification=raggedright,     
}
\caption{An example of the generated sentences by our proposed system for a test scenario from RoboCup dataset.}
\label{table4}
\fontsize{8}{18}\selectfont
\begin{tabular}{>{\centering\arraybackslash} m{2cm}|>{\centering\arraybackslash} m{13cm}} 
\hline
\multicolumn{1}{l|}{\textbf{MR}} & \multicolumn{1}{l}{\vtop{\hbox{\strut \textbf{\{type: ”defense", @arg1: ”pink4", @arg2: "pink4"}\}}\hbox{\strut \textbf{\{type: ”turnover", @arg1: ”purple7", @arg2: "pink4"}\}}}} \\ \hline
\multicolumn{1}{l|}{\textbf{Reference}}  & \multicolumn{1}{l}{purple7 turned the ball over to pink4} \\ \hline
\multirow{10}[2]{*}{\vtop{\hbox{\strut \textbf{The generated}}\hbox{\strut \textbf{sentences}}} }
& \multicolumn{1}{l} {purple7 turned the ball over to pink4}\\
& \multicolumn{1}{l}{purple7 loses the ball over to pink4}\\
& \multicolumn{1}{l}{purple7 started things by turning over the ball to pink4}\\
& \multicolumn{1}{l}{purple7 immediately turns the ball over to pink4}\\
& \multicolumn{1}{l}{purple7 loses the ball back over to pink4}\\
& \multicolumn{1}{l}{purple7 immediately turns the ball back over to pink4}\\
& \multicolumn{1}{l}{purple7 immediately turns the ball over to pink4}\\
& \multicolumn{1}{l}{purple7 lost the ball and turned it over to pink4}\\ 
& \multicolumn{1}{l}{purple7 turned the ball and turned it over to pink4}\\ \hline
\end{tabular}
\end{center}
\end{table}
\begin{table}
\begin{center}
\captionsetup{%
    font={small}, 
    format=plain,  
    singlelinecheck=true,      
    labelsep=newline,
    labelfont=bf,      
}
\caption{Results of Human Evaluations on three tabular format datasets in terms of Informativeness, Naturalness and Quality, rating out of 5. For baseline we used gold-standard references in each dataset.}
\label{table5}
\fontsize{8.5}{12}\selectfont
\begin{tabular}{+l^c^c^c^c^c^c}
\hline
\rowstyle{\bfseries}
\multirow{2}[4]{*}{Dataset} & \multicolumn{2}{c}{\textbf{Informativeness}} &
\multicolumn{2}{c}{\textbf{Naturalness}} & \multicolumn{2}{c}{\textbf{Quality}} \\
\cmidrule(lr){2-3} \cmidrule(lr){4-5} \cmidrule(lr){6-7}
 & \multicolumn{1}{c}{\textbf{Ours}}
 & \multicolumn{1}{c}{\textbf{Baseline}}
 & \multicolumn{1}{c}{\textbf{Ours}}
 & \multicolumn{1}{c}{\textbf{Baseline}}
  & \multicolumn{1}{c}{\textbf{Ours}}
 & \multicolumn{1}{c}{\textbf{Baseline}} \\
 \hline
Atis & 5.00 & 5.00 & 4.09 & 3.97 & 4.59 & 4.57 \\
RoboCup&4.75 & 5.00 & 4.12 & 4.08 & 4.85 & 4.76 \\
WEATHERGOV & 5.00 & 5.00 & 4.60 & 4.53 &4.78 & 4.78 \\
 \hline
\end{tabular}
\end{center}
\end{table}
\subsubsection{General domain experiment}
\begin{figure}
\captionsetup{%
    font={small}, 
    format=plain,  
    singlelinecheck=false,      
    labelsep=newline,
    labelfont=bf,      
}
\center
\includegraphics[width=1\textwidth]{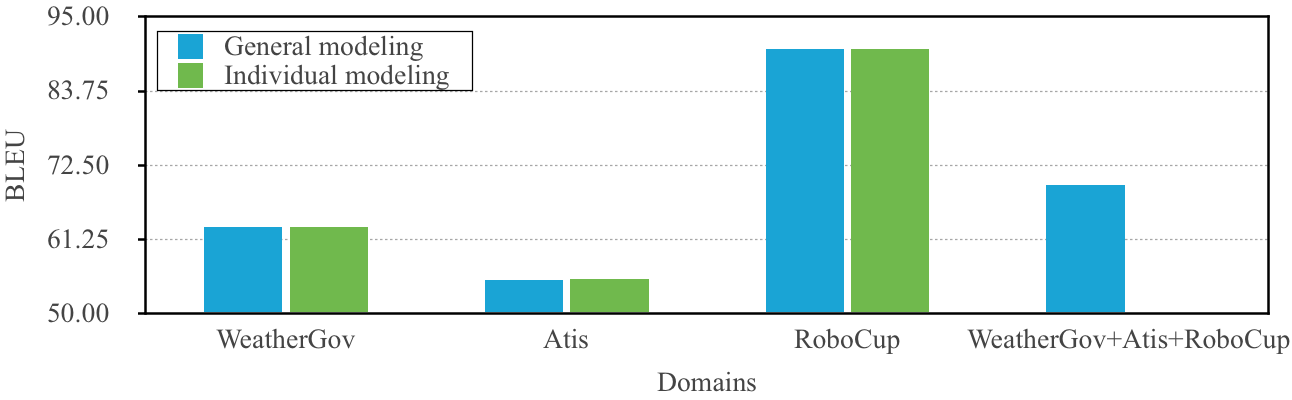} 
\caption{Performance of the general model on pooling three different tabular format datasets.}
\label{fig: Figure7}
\end{figure}

In this experiment, we trained our system by pooling all the data from three different domains together and examined them in each domain. The performance comparison is shown in Figure \ref{fig: Figure7}. As can be seen, in most of the datasets, the BLEU scores of individual training and general training are not different (the difference is less than 0.2 units out of 100); since each dataset has a different set of meaning labels and uses different sets of words based on the concept of its domain. These results show that our proposed NLG system can work simultaneously with data from different conceptual domains.

\subsection{Experiments on dialogue act datasets}
\subsubsection{Experimental Setup}
As mentioned before, for extracting dependency tree from the aligned training sentences, we used CoreNLP dependency parser provided by Stanford University. For each test scenario, we over-generated 20 sentences ($B=20$, $C=20$) and selected the 5 ones with the most ‌scores following \namecite{WenA,WenB,WenC,WenD,TranA,TranB}. We used the corpus-based-BLEU-4 metric for the objective evaluation by adopting code from an open-source benchmark toolkit for Natural Language Generation\footnote{ https://github.com/shawnwun/RNNLG}. Also, we used the slot error rate (ERR), the rate of slots that are generated redundantly or missing in the generated sentence, and is computed as follows:
\begin{equation}
\label{e7}
ERR=\frac {r+m} {N}
\end{equation}
where $r$ and $m$ are the numbers of redundant or missing slot and $N$ is the total number of slots in the dialogue act of the test scenario. For each domain, we generated multiple references from the sentences in the validation set with similar meaning labels, following \namecite{WenA}. Moreover, we also ran human evaluation to evaluate the generated sentences by our proposed system in terms of informativeness, naturalness, and quality. To do this, we selected randomly 20 test scenarios for each dataset. And like before, we used 10 students from the University of Amsterdam, whose native language was English (Fleiss’s $\kappa$ =0.69, Krippendorff’s $\alpha$ =0.72), for judging the generated sentences, without knowing each sentence is produced by what system. We compared our proposed model against three models released from the NLG toolkit such as ENCDEC \cite{WenC}, HLSTM  \cite{WenA}, and SCLSTM  \cite{WenB,WenD} and also SRGRU-Context, ESRGRU-MUL, ESRGRU-INNER and RALSTM from RNN gating enhancement \cite{TranA,TranB} that are trained on Hotel, Restaurant, TV and Laptop dialogue act datasets.

\subsubsection{Individual domain experiment}
In this experiment, again, we trained our proposed NLG system on each dataset separately and then compared it against baseline systems. Table \ref{table6} shows the objective evaluation results. As can be seen, our system achieves a comparable result with the baseline systems in terms of both BLEU and ERR. We achieved zero ERR because in the content planning step of our proposed system, we considered all the selected meaning labels of a test dialogue act. As a result, in the surface realization step, the produced dependency trees are selected in a way that contain all the required meaning labels. To illustrate the ability of our proposed system to generate a set of diverse sentences related to the same concept, we showed a few of the generated sentences for a test scenario from Hotel dataset in Table \ref{table7}. 

Tables \ref{table8} and \ref{table9} show the mean scores of each human evaluation factor. Since in our proposed system all required meaning labels are forced to be in the produced dependency tree, the mean score for informativeness factor is 5. Furthermore, as can be seen, the sentences generated by our system received high scores from the judges for the naturalness and quality factors.

\begin{table}
\begin{center}
\captionsetup{%
    font={small}, 
    format=plain,  
    singlelinecheck=true,      
    labelsep=newline,
    labelfont=bf,      
}
\caption{Performance of different models on four dialogue act format datasets in terms of the BLEU and ERR scores; bold denotes the best model\textsuperscript{*}.}
\label{table6}
\fontsize{7}{13}\selectfont
\begin{tabular}{+l^r^r^r^r^r^r^r^r}
\hline
\rowstyle{\bfseries}
\multirow{2}[3]{*}{Model} & \multicolumn{2}{c}{\textbf{Restaurant}} & \multicolumn{2}{c}{\textbf{Hotel}} & \multicolumn{2}{c}{\textbf{TV}} & \multicolumn{2}{c}{\textbf{Laptop}} \\
\cmidrule(lr){2-3} \cmidrule(lr){4-5}\cmidrule(lr){6-7}\cmidrule(lr){8-9}
 & \multicolumn{1}{c}{\textbf{BLEU}}
 & \multicolumn{1}{c}{\textbf{ERR(\%)}}
 & \multicolumn{1}{c}{\textbf{BLEU}}
 & \multicolumn{1}{c}{\textbf{ERR(\%)}}
 & \multicolumn{1}{c}{\textbf{BLEU}}
 & \multicolumn{1}{c}{\textbf{ERR(\%)}}
 & \multicolumn{1}{c}{\textbf{BLEU}}
 & \multicolumn{1}{c}{\textbf{ERR(\%)}} \\
 \hline
SCLSTM & 75.25 & 0.38 & 84.82 & 3.07 & 52.65 & 2.31 & 51.16 & 0.79 \\
HLSTM & 74.66 & 0.74 & 85.04 & 2.67 & 52.50 & 2.50 & 51.34 & 1.10 \\
ENCDEC & 73.98 & 2.78 & 85.49 & 4.69 & 51.82 & 3.18 & 51.08 & 4.04\\
SRGRU-Context & 76.34 & 0.94 & 87.76 & 0.98 & 53.11 & 1.33 & 51.19 & 1.19\\ 
ESRGRU-MUL & 76.49 & 1.01 & 88.99 & 0.53 & 53.21 & 0.90 & 52.23 &	1.10\\ 
ESRGRU-INNER & 76.56 & 0.76& 89.67 & 0.94 & 53.30 & 0.90 & 52.36 & 0.90 \\ 
RALSTM & 77.89 & 0.16 & 89.81 & 0.43 & 54.06 & 0.63 & 52.52 & 0.42 \\ \hline
\textbf{Our System} & \textbf{77.89} & \textbf{0.00} & \textbf{90.21} & \textbf{0.00} & \textbf{56.39} & \textbf{0.00} & \textbf{55.18} & \textbf{0.00}\\  \hline
  \multicolumn{9}{p{0.85\textwidth}}{\textsuperscript{*}\small \textit{Notes. The BLEU and ERR scores for baseline systems are reported scores by their authors.}}
\end{tabular}
\end{center}
\end{table}
\begin{table}
\begin{center}
\captionsetup{%
    font={small}, 
    format=plain,  
    singlelinecheck=false,      
    labelsep=newline,
    labelfont=bf, 
    justification=raggedright,     
}
\caption{An example of the generated sentences by our proposed system for a test scenario from Hotel dataset.}
\label{table7}
\fontsize{8}{18}\selectfont
\begin{tabular}{>{\centering\arraybackslash} m{2cm}|>{\centering\arraybackslash} m{13cm}} 
\hline
\multicolumn{1}{l|}{\textbf{MR}} & \multicolumn{1}{l}{ inform(name='grant hotel';pricerange='inexpensive';dogsallowed='no')} \\ \hline
\multicolumn{1}{l|}{\textbf{Reference}}  & \multicolumn{1}{l}{the grant hotel does not allow dogs and is inexpensive} \\ \hline
\multirow{10}[2]{*}{\vtop{\hbox{\strut \textbf{The generated}}\hbox{\strut \textbf{sentences}}} }
& \multicolumn{1}{l} {the grant hotel does not allow dogs and is in the inexpensive price range}\\
& \multicolumn{1}{l}{grant hotel is an inexpensive hotel that does not allow dogs}\\
& \multicolumn{1}{l}{the grant hotel does not allow dogs and is pretty inexpensive}\\
& \multicolumn{1}{l}{yes the grant hotel does not allow dogs and in the inexpensive price range}\\
& \multicolumn{1}{l}{a hotel that does not allow dogs is the inexpensive priced grant hotel}\\
& \multicolumn{1}{l}{the grant hotel does not allow dogs however it is inexpensive}\\
& \multicolumn{1}{l}{i found the grant hotel in the inexpensive price range that does not allow dogs}\\
& \multicolumn{1}{l}{okay well the grant hotel in the inexpensive price range that does not allow dogs}\\ 
& \multicolumn{1}{l}{great there is also the grant hotel in the inexpensive price range that does not allow dogs}\\
& \multicolumn{1}{l}{okay i found the grant hotel with an inexpensive price range that does not allow dogs}\\ \hline
\end{tabular}
\end{center}
\end{table}
\begin{table}
\begin{center}
\captionsetup{%
    font={small}, 
    format=plain,  
    singlelinecheck=true,      
    labelsep=newline,
    labelfont=bf,      
}
\caption{Results of Human Evaluations on Hotel and Restaurant datasets in terms of Informativeness, Naturalness and Quality, rating out of 5. For baselines, the output sentences are generated by training each network on 5 random initialization\textsuperscript{*}.}
\label{table8}
\fontsize{8.5}{12}\selectfont
\begin{tabular}{+l^c^c^c^c^c^c}
\hline
\rowstyle{\bfseries}
\multirow{2}[4]{*}{Model} & \multicolumn{2}{c}{\textbf{Informativeness}} &
\multicolumn{2}{c}{\textbf{Naturalness}} & \multicolumn{2}{c}{\textbf{Quality}} \\
\cmidrule(lr){2-3} \cmidrule(lr){4-5} \cmidrule(lr){6-7}
 & \multicolumn{1}{c}{\textbf{Hotel}}
 & \multicolumn{1}{c}{\textbf{Restaurant}}
 & \multicolumn{1}{c}{\textbf{Hotel}}
 & \multicolumn{1}{c}{\textbf{Restaurant}}
  & \multicolumn{1}{c}{\textbf{Hotel}}
 & \multicolumn{1}{c}{\textbf{Restaurant}} \\
 \hline
SCLSTM & 4.73 & 4.96 & 4.11 & 3.54 & 4.17 & 4.13 \\
HLSTM&4.80 & 4.90 & 3.97 & 3.25 & 4.39 & 3.74\\
 ENCDEC & 4.65 & 4.78 & 4.03 & 3.06 &4.47 & 4.11 \\
SRGRU-Context & 4.91 & 4.85 & 4.09 & 3.42 & 4.42 & 4.18 \\ 
ESRGRU-MUL & 4.95 & 4.81 & 4.06 & 3.48 & 4.38 & 4.36 \\ 
ESRGRU-INNER & 4.93 & 4.89 & 4.23 & 3.72 & 4.68 & 4.04\\ 
RALSTM & 4.92 & 4.95 & 4.07 & 4.50 & 4.71 & 4.58\\ 
Our System & 5.00 & 5.00 & 4.50 & 4.08 & 4.85 & 4.77\\
 \hline
  \multicolumn{7}{p{0.9\textwidth}}{\textsuperscript{*}\small \textit{Notes. For baseline systems, we generated the output sentences using the RNNLG toolkit and our implementation of the RNN gating enhancement papers.}} 
\end{tabular}
\end{center}
\end{table}
\begin{table}
\begin{center}
\captionsetup{%
    font={small}, 
    format=plain,  
    singlelinecheck=true,      
    labelsep=newline,
    labelfont=bf,      
}
\caption{Results of Human Evaluations on TV and Laptop datasets in terms of Informativeness, Naturalness and Quality, rating out of 5. For baselines, the output sentences are generated by training each network on 5 random initialization\textsuperscript{*}.}
\label{table9}
\fontsize{8.5}{12}\selectfont
\begin{tabular}{+l^c^c^c^c^c^c}
\hline
\rowstyle{\bfseries}
\multirow{2}[4]{*}{Model} & \multicolumn{2}{c}{\textbf{Informativeness}} &
\multicolumn{2}{c}{\textbf{Naturalness}} & \multicolumn{2}{c}{\textbf{Quality}} \\
\cmidrule(lr){2-3} \cmidrule(lr){4-5} \cmidrule(lr){6-7}
 & \multicolumn{1}{c}{\textbf{TV}}
 & \multicolumn{1}{c}{\textbf{Laptop}}
 & \multicolumn{1}{c}{\textbf{TV}}
 & \multicolumn{1}{c}{\textbf{Laptop}}
  & \multicolumn{1}{c}{\textbf{TV}}
 & \multicolumn{1}{c}{\textbf{Laptop}} \\
 \hline
SCLSTM & 4.61 & 4.53 & 3.89 & 3.93 & 4.57 & 4.37 \\
HLSTM&4.55 & 4.00 & 4.06 & 4.01 & 4.56 & 4.47\\
 ENCDEC & 4.51 & 4.10 & 4.23 & 4.05 &4.67 & 4.51 \\
SRGRU-Context & 4.70 & 4.38 & 4.30 & 4.19 & 4.55 & 4.63 \\ 
ESRGRU-MUL & 4.72 & 4.43 & 4.38 & 4.29& 4.52 & 4.77 \\ 
ESRGRU-INNER & 4.72 & 4.44 & 4.41 & 4.30 & 4.59 & 4.72\\ 
RALSTM & 4.81 & 4.73 & 4.53 & 4.49 & 4.73 & 4.85\\ 
Our System & 5.00 & 5.00 & 4.65 & 4.51 & 4.81 & 4.85\\
 \hline
  \multicolumn{7}{p{0.8\textwidth}}{\textsuperscript{*}\small \textit{Notes. For baseline systems, we generated the output sentences using the RNNLG toolkit and our implementation of the RNN gating enhancement papers.}} 
\end{tabular}
\end{center}
\end{table}
\begin{table}
\begin{center}
\captionsetup{%
    font={small}, 
    format=plain,  
    singlelinecheck=true,      
    labelsep=newline,
    labelfont=bf,      
}
\caption{Performance of different models on pooling Hotel and Restaurant datasets in term of BLEU score. For baselines, the results are produced by training each network on 5 random initialization\textsuperscript{*}.}
\label{table10}
\fontsize{8.5}{12}\selectfont
\begin{tabular}{+l^c^c^c^c^c^c}
\hline
\rowstyle{\bfseries}
\multirow{2}[5]{*}{Model} & \multicolumn{2}{c}{\textbf{Hotel}} &
\multicolumn{2}{c}{\textbf{Restaurant}} & \multirow{2}[5]{*}{Hotel+Restaurant} \\
\cmidrule(lr){2-3} \cmidrule(lr){4-5} 
 & \multicolumn{1}{c}{\vtop{\hbox{\strut \textbf{Individual}}\hbox{\strut \textbf{model}}} }
 & \multicolumn{1}{c}{\vtop{\hbox{\strut \textbf{General}}\hbox{\strut \textbf{model}}} }
 & \multicolumn{1}{c}{\vtop{\hbox{\strut \textbf{Individual}}\hbox{\strut \textbf{model}}} }
 & \multicolumn{1}{c}{\vtop{\hbox{\strut \textbf{General}}\hbox{\strut \textbf{model}}} } \\
 \hline
SCLSTM & 84.82 & 82.67 & 75.25 & 71.80 & 80.80 \\
HLSTM&85.04 & 81.74 & 74.66 & 66.50 & 81.00 \\
 ENCDEC & 85.49 & 84.09 & 73.98 & 74.70 & 83.22 \\
SRGRU-Context & 87.76 & 85.70 & 76.34 & 72.12 & 85.05 \\ 
ESRGRU-MUL & 88.99 & 88.00 & 76.49 & 73.87 & 86.92 \\ 
ESRGRU-INNER & 89.76 & 88.37 & 76.56 & 74.19 & 87.03\\ 
RALSTM & 89.81 & 88.81 & 77.89 & 74.5 & 87.76\\ 
Our System & 90.21 & 89.79 & 77.89 & 75.39 & 87.91\\
 \hline
  \multicolumn{7}{p{0.9\textwidth}}{\textsuperscript{*}\small \textit{Notes. For baseline systems, we generated the output sentences using the RNNLG toolkit and our implementation of the RNN gating enhancement papers.}}  
\end{tabular}
\end{center}

\end{table}
\begin{table}
\begin{center}
\captionsetup{%
    font={small}, 
    format=plain,  
    singlelinecheck=true,      
    labelsep=newline,
    labelfont=bf,      
}
\caption{Performance of different models on pooling TV and Laptop datasets in term of BLEU score. For baselines, the results are produced by training each network on 5 random initialization\textsuperscript{*}.}
\label{table11}
\fontsize{8.5}{12}\selectfont
\begin{tabular}{+l^c^c^c^c^c^c}
\hline
\rowstyle{\bfseries}
\multirow{2}[5]{*}{Model} & \multicolumn{2}{c}{\textbf{TV}} &
\multicolumn{2}{c}{\textbf{Laptop}} & \multirow{2}[5]{*}{TV+Laptop} \\
\cmidrule(lr){2-3} \cmidrule(lr){4-5} 
 & \multicolumn{1}{c}{\vtop{\hbox{\strut \textbf{Individual}}\hbox{\strut \textbf{model}}} }
 & \multicolumn{1}{c}{\vtop{\hbox{\strut \textbf{General}}\hbox{\strut \textbf{model}}} }
 & \multicolumn{1}{c}{\vtop{\hbox{\strut \textbf{Individual}}\hbox{\strut \textbf{model}}} }
 & \multicolumn{1}{c}{\vtop{\hbox{\strut \textbf{General}}\hbox{\strut \textbf{model}}} } \\
 \hline
SCLSTM & 52.65 & 50.11 & 51.16 & 49.32 & 48.8 \\
HLSTM & 52.5 & 49.61 & 51.34 & 50.15 & 48.04 \\
 ENCDEC & 51.82 & 48.7 & 51.08 & 49.05 & 47.11 \\
SRGRU-Context &53.11 & 50.46 & 51.14 & 49.4 & 48.63 \\ 
ESRGRU-MUL & 53.21 & 49.7 & 52.23 & 50.13 & 49.01 \\ 
ESRGRU-INNER & 53.30 & 50.2 & 52.36 & 50.53 & 49.42\\ 
RALSTM & 54.06 & 51.5 & 52.52 & 50.87 & 49.9\\ 
Our System & 56.39 & 52.12 & 55.18 & 52.54 & 51.14\\
 \hline
  \multicolumn{7}{p{0.9\textwidth}}{\textsuperscript{*}\small \textit{Notes. For baseline systems, we generated the output sentences using the RNNLG toolkit and our implementation of the RNN gating enhancement papers.}}  
\end{tabular}
\end{center}
\end{table}

\subsubsection{General domain experiment}
In this experimental comparison, again, we trained our system by pooling all the data from two different domains, with the most shared meaning labels together and examined them in individual domains as well as the pooled domain. The performance comparison is shown in Table \ref{table10} and \ref{table11} for pooling the Hotel and Restaurant domains and also pooling the TV and Laptop domains. As can be seen, after pooling, for all models the BLEU scores are decreased a little compared to when they were trained on individual domains. This was not unexpected. Clearly, both the Hotel and Restaurant domains are about booking a place, as well as TV and Laptop domains are about buying an electronic device, so they have many common meaning labels and words. But for the Hotel and Laptop domains, the number of training and validation examples are larger and their sentences are longer (Table \ref{table1}). Accordingly, for the dialogue acts that have the same act type and slots in the both pooled domains, the length of generated sentences and words are used in the them can be more biased to the Hotel and Laptop domains. As a result, as can be seen in Table \ref{table10} and \ref{table11}, the difference in BLEU scores between the individual and general model for the Hotel and Laptop domains are less than the Restaurant and TV domains.

\subsubsection{Unseen domain experiment}
\begin{figure}
    \centering
    \begin{subfigure}[b]{\textwidth}
        \includegraphics[width=\textwidth]{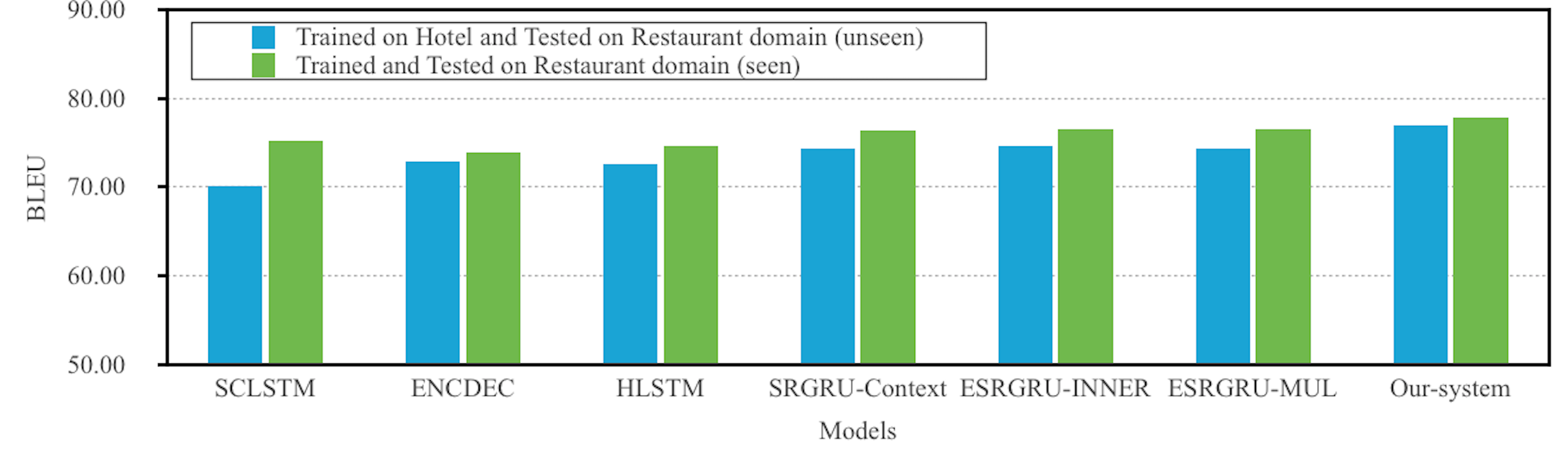} 
        \caption{BLEU comparisons}
        \label{fig:B}
    \end{subfigure}
    \begin{subfigure}[b]{\textwidth}
         \includegraphics[width=\textwidth]{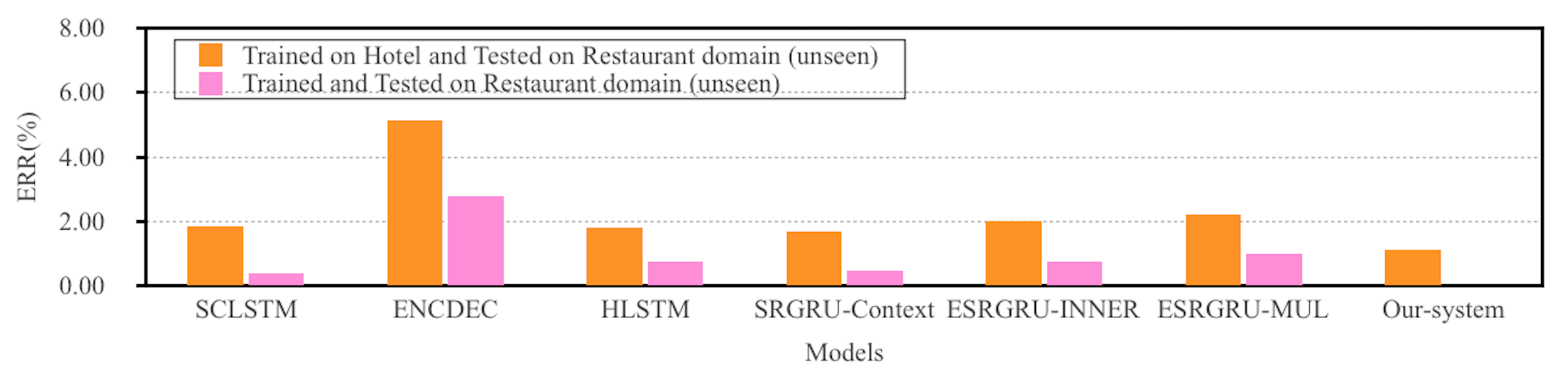}
        \caption{ERR comparisons}
        \label{fig:E}
    \end{subfigure}
\captionsetup{%
    font={small}, 
    format=plain,  
    singlelinecheck=true,      
    labelsep=newline,
    labelfont=bf,      
}
\caption{ Performance of all models trained with Hotel dataset and tested on Restaurant as unseen dataset.}   
\label{fig: figure7} 
\end{figure}

\begin{figure}
    \centering
    \begin{subfigure}[b]{\textwidth}
        \includegraphics[width=\textwidth]{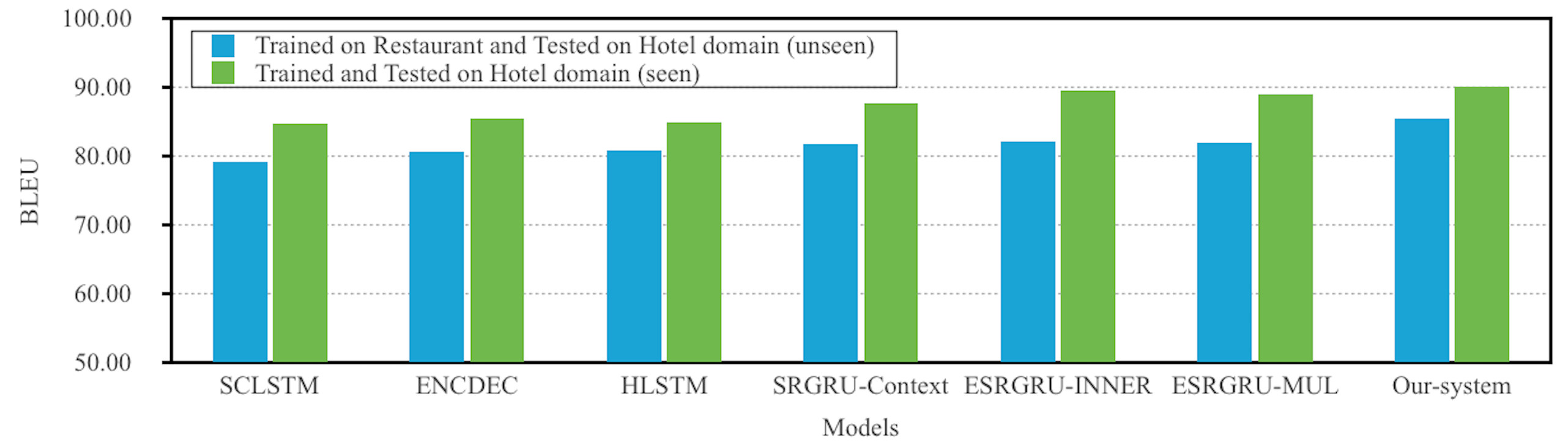} 
        \caption{BLEU comparisons}
        \label{fig:B}
    \end{subfigure}
    \begin{subfigure}[b]{\textwidth}
         \includegraphics[width=\textwidth]{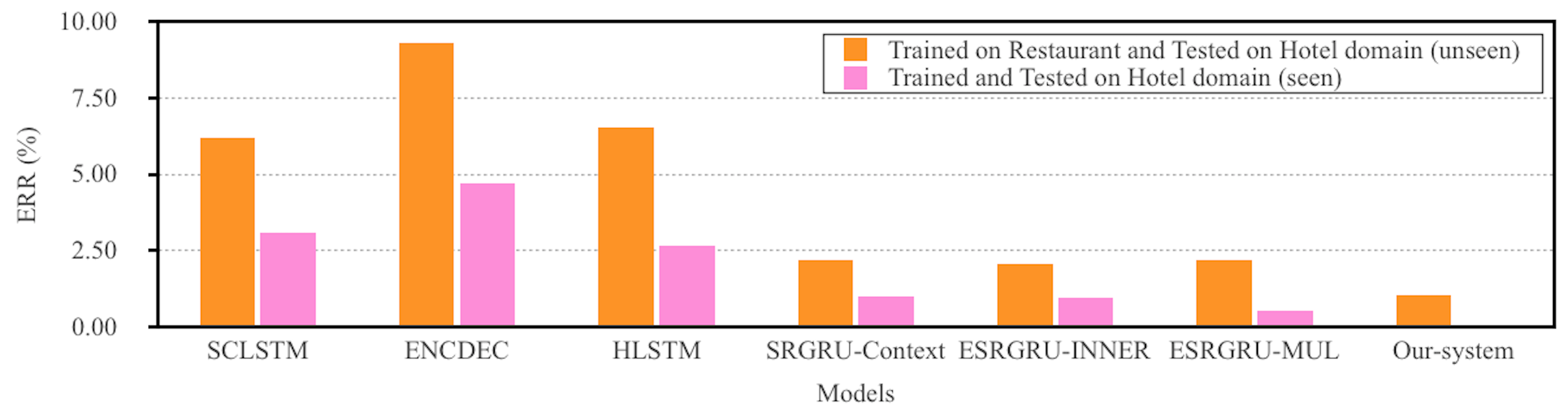}
        \caption{ERR comparisons}
        \label{fig:E}
    \end{subfigure}
\captionsetup{%
    font={small}, 
    format=plain,  
    singlelinecheck=true,      
    labelsep=newline,
    labelfont=bf,      
}
\caption{ Performance of all models trained with Restaurant dataset and tested on Hotel as unseen dataset.}   
\label{fig: figure8} 
\end{figure}

\begin{figure}
    \centering
    \begin{subfigure}[b]{\textwidth}
        \includegraphics[width=\textwidth]{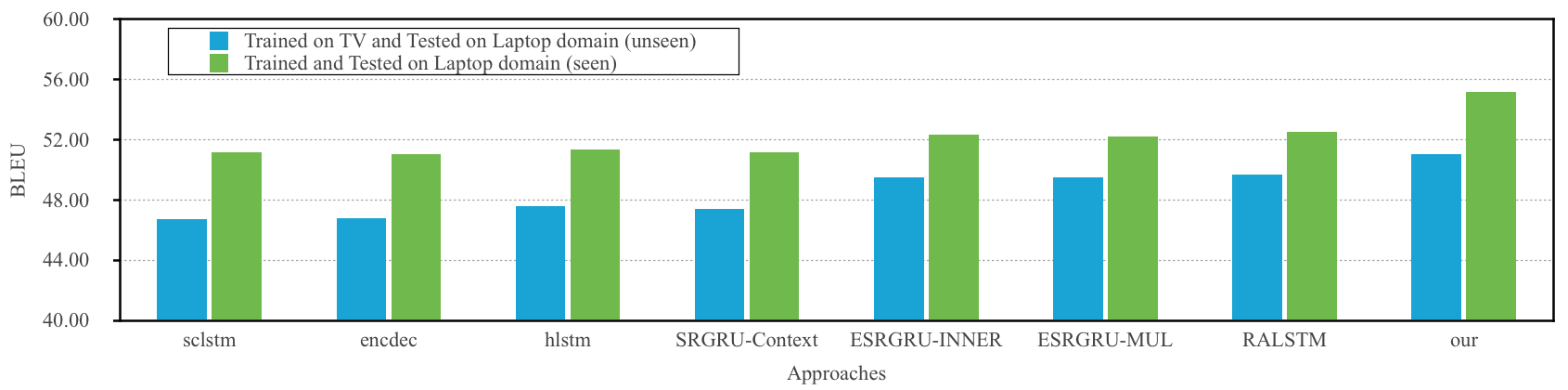} 
        \caption{BLEU comparisons}
        \label{fig:B}
    \end{subfigure}
    \begin{subfigure}[b]{\textwidth}
         \includegraphics[width=\textwidth]{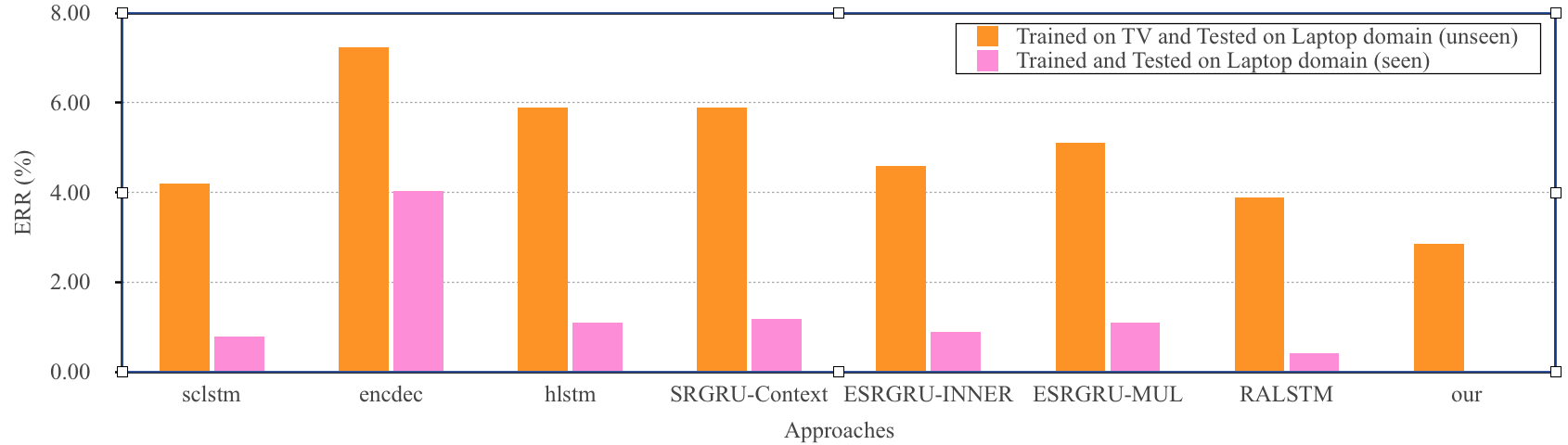}
        \caption{ERR comparisons}
        \label{fig:E}
    \end{subfigure}
\captionsetup{%
    font={small}, 
    format=plain,  
    singlelinecheck=true,      
    labelsep=newline,
    labelfont=bf,      
}
\caption{ Performance of all models trained with TV dataset and tested on Laptop as unseen dataset.}   
\label{fig: figure9} 
\end{figure}

\begin{figure}
    \centering
    \begin{subfigure}[b]{\textwidth}
        \includegraphics[width=\textwidth]{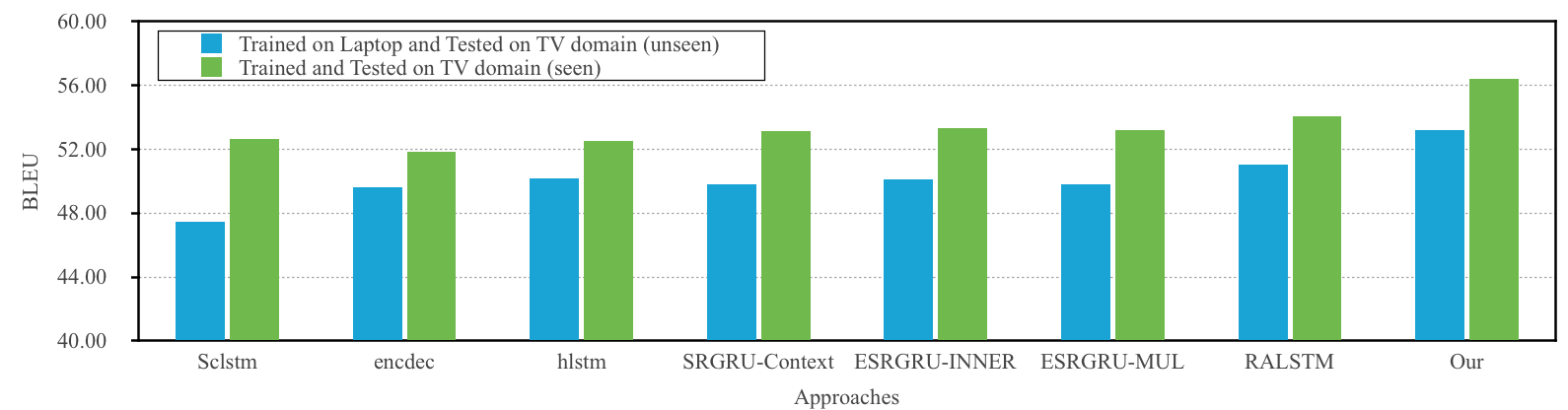} 
        \caption{BLEU comparisons}
        \label{fig:B}
    \end{subfigure}
    \begin{subfigure}[b]{\textwidth}
         \includegraphics[width=\textwidth]{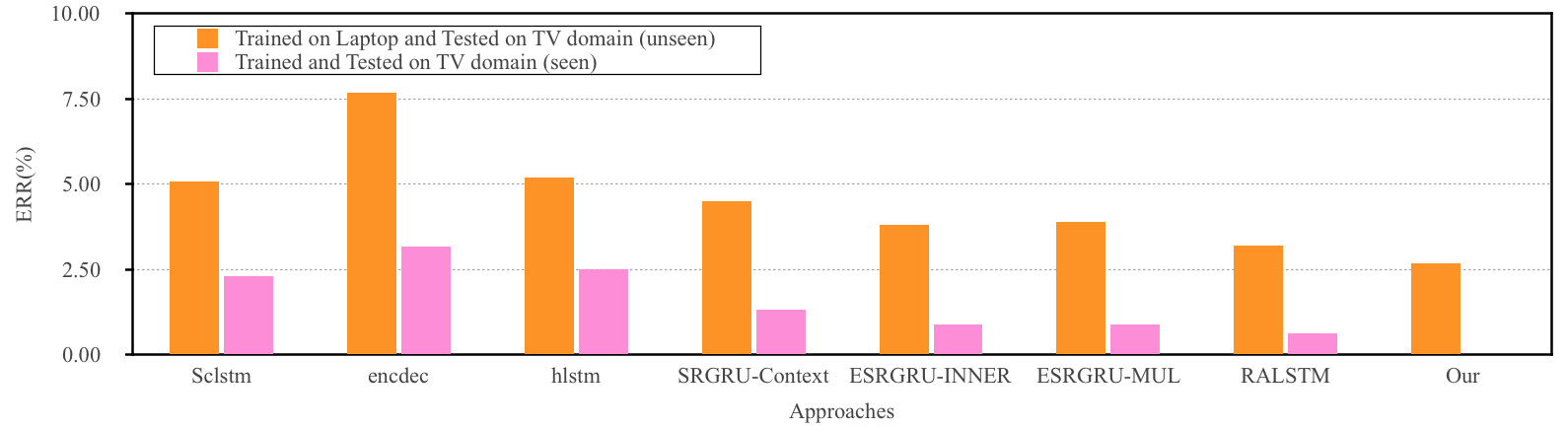}
        \caption{ERR comparisons}
        \label{fig:E}
    \end{subfigure}
\captionsetup{%
    font={small}, 
    format=plain,  
    singlelinecheck=true,      
    labelsep=newline,
    labelfont=bf,      
}
\caption{ Performance of all models trained with Laptop dataset and tested on TV as unseen dataset.}   
\label{fig: figure10} 
\end{figure}

In this experiment\footnote{For baseline systems, we generated the output sentences using the RNNLG toolkit and our implementation of the RNN gating enhancement papers.}, we tested the performance of our proposed system for unseen data against other baselines models. Due to the shared meaning labels between the Hotel and Restaurant domains, as well as TV and Laptop domains, we performed this test individually for these two sets of data. To do this, one domain of each set of data was selected as training data and the other as unseen data. Also, in our system, we relaxed the constraints on mentioning all the meaning labels of the test scenario in the generated sentences. The performance comparisons are shown in Figures \ref{fig: figure7}, \ref{fig: figure8}, \ref{fig: figure9} and \ref{fig: figure10}. As can be seen, for both sets of data and for all models, the BLEU scores decreased and ERR scores increased. These differences in both scores are larger when the Hotel and Laptop domains are used as unseen domain; due to the different distributions of un-shared meaning labels in the test scenarios of each domain.

\begin{table}
\begin{center}
\captionsetup{%
    font={small}, 
    format=plain,  
    singlelinecheck=true,      
    labelsep=newline,
    labelfont=bf,      
}
\caption{Performance of the proposed system on E2E dataset based on different Beam widths (B) and number of candidate sentences (C) in terms of the BLEU, NIST, METEOR, ROUGE-L and CIDEr scores.}
\label{table12}
\fontsize{8.5}{15}\selectfont
\begin{tabular}{+l ^c ^c ^c^c^c}
\hline
\rowstyle{\bfseries}
Metric & B=1, C=1 & B=5, C=5 & B=10, C=10 & B=15, C=15 & B=20, C=20\\ 
 \hline
BLEU & 40.16 & 51.95 & 58.23 & 62.78 & 66.05\\
METEOR & 0.3973 & 0.4216 & 0.4313 & 0.4403& 0.447\\
NIST & 6.27 & 6.91 & 7.25 & 8.01 & 8.61\\
ROUGE-L& 0.5187 & 0.5511 & 0.5962 & 0.6111 & 0.6751\\
CIDEr& 1.33 & 1.62 & 1.85 & 2.01 & 2.19\\
 \hline
\end{tabular}
\end{center}
\end{table}

\begin{table}
\begin{center}
\captionsetup{%
    font={small}, 
    format=plain,  
    singlelinecheck=true,      
    labelsep=newline,
    labelfont=bf,      
}
\caption{Performance of different models on E2E dataset in terms of the BLEU, NIST, METEOR, ROUGE-L and CIDEr metrics; bold denotes the best scores\textsuperscript{*}.}
\label{table13}
\fontsize{7.5}{15}\selectfont
\begin{tabular}{>{\centering\arraybackslash}{l} >{\centering\arraybackslash}{c} >{\centering\arraybackslash}{c} >{\centering\arraybackslash}{c} >{\centering\arraybackslash}{c} >{\centering\arraybackslash}{c}}
\hline
\rowstyle{\bfseries}
 Methods &  BLEU  &  METEOR & NIST  &  ROUGE-L &  CIDEr  \\ \hline
Baseline\cite{Dusek} & 65.93 & \textbf{0.4483} & \textbf{8.61} & 0.6850 & 2.23 \\
SLUG\cite{Juraska} & \textbf{66.19} & 0.4454 & \textbf{8.61} & 0.6772 & 2.26 \\
HARV\cite{Gehrmann} & 64.96 & 0.4386 & 8.53 & 0.6872 & 2.09 \\
Zhang\cite{Zhang} & 65.45 & 0.4392 & 8.18 & \textbf{0.7083} & 2.10 \\
Gong\cite{Gong} & 64.22 & 0.4469 & 	8.35 & 0.6645 & \textbf{2.27}\\ 
Our system & 66.05 & 0.447 & \textbf{8.61} &	0.6751 & 2.19\\ 
 \hline
  \multicolumn{6}{p{0.8\textwidth}}{\textsuperscript{*}\small \textit{Notes. The scores for baseline and other systems are reported scores by their authors.}} 
\end{tabular}
\end{center}
\end{table}
\begin{table}
\begin{center}
\captionsetup{%
    font={small}, 
    format=plain,  
    singlelinecheck=true,      
    labelsep=newline,
    labelfont=bf,      
}
\caption{Results of Human Evaluations on E2E datasets in terms of Informativeness, Naturalness and Quality, rating out of 5. For baselines, the reported sentences by challenge organizers are used.}
\label{table14}
\fontsize{7.5}{15}\selectfont
\begin{tabular}{>{\centering\arraybackslash}{l} >{\centering\arraybackslash}{c} >{\centering\arraybackslash}{c} >{\centering\arraybackslash}{c}}
\hline
\rowstyle{\bfseries}
 Methods &  Informativeness  &  Naturalness & Quality   \\ \hline
Baseline & 4.51 &  4.19 & 4.41 \\
SLUG  & 4.63 &  4.47 & 4.83 \\
HARV & 4.59 & 4.43 & 4.87  \\
Zhang& 4.57 & 4.35 & 4.61 \\
Gong & 4.48 & 	4.34 & 4.86 \\ 
Our system & 5.00 & 4.52 & 4.87 \\ 
 \hline
\end{tabular}
\end{center}
\end{table}
\subsection{Experiments on E2E dataset}
Here also we used CoreNLP dependency parser for extracting dependency tree from the aligned training sentences. For generating dependency trees for each test scenario, we used Beam search width $B={\{1,5,10,15,20}\}$ and after generating sentences from the trees, for each $B$, we selected the most probable sentence as the output. For objective evaluation, the BLEU-4, NIST, METEOR, ROUGE-L, and CIDEr metrics are used by the provided evaluation code for E2E challenge\footnote{ https://github.com/tuetschek/e2e-metrics}. We also ran human evaluation in order to evaluate the generated sentences by our proposed system in terms of the informativeness, naturalness and quality. To do this, we selected randomly 20 test scenarios. The judges were 10 students from the University of Amsterdam, whose native language was English (Fleiss’s $\kappa$ =0.69, Krippendorff’s $\alpha$ =0.72). They were not informed about the system that had produced a specific sentence. We compared our proposed model against baseline \cite{Dusek} and the best four methods that participate in the E2E challenge \cite{Juraska,Gehrmann,Zhang,Gong}. Unlike other systems that performed pre-processing on input data or post-processing on output sentences to provide better quality, we did not change the input or output because we wanted to maintain the generality of our system. The objective evaluation results for different $B$ values are shown in Table \ref{table12} and overall comparisons are shown in Table \ref{table13}. Due to the nature of the proposed method, in $B=1$ which is actually a greedy search, the scores are low and by using larger values for $B$ the scores are also increased. Also, our system achieved a better result than the baseline system. As mentioned earlier, most of the systems submitted in this challenge were sequence-to-sequence based systems. The best result was for \namecite{Juraska} that used an Encoder and 3 different Decoders, so the output sentences were selected by choosing the best-generated sentence from all Decoders. Although our proposed system did not outperform this method, it achieved a comparable result against other submitted methods. Table \ref{table14} shows the mean scores of each human evaluation factor. As can be seen, the sentences generated by our system received high scores from the judges for the informativeness, naturalness and quality factors.

\subsection{Experiments on WebNLG dataset}
Like before, for extracting dependency tree from the aligned training sentences, we used CoreNLP dependency parser. For generating dependency trees for each test scenario, we used Beam search width $B={\{1,5,10,15,20}\}$ and after generating sentences from the trees, for each $B$, we selected the most probable sentence as the output. Also, for the objective evaluation, we used the BLEU-4, METEOR and TER metrics that were determined by WebNLG challenge organizers \citep{Gardent}. We also ran human evaluation in order to evaluate the generated sentences by our proposed system in terms of naturalness and quality. To do this, we selected randomly 20 test scenarios. And like before, we used 10 students from the University of Amsterdam, whose native language was English (Fleiss’s $\kappa$ =0.69, Krippendorff’s $\alpha$ =0.72), for judging the generated sentences, without knowing each sentence is produced by what system. We compared our proposed model against baseline and the best four methods that participate in this challenge, in three ways. First, we trained and tested our proposed system on 10 seen domains from all 15 domains in WebNLG dataset, that their train and validation data were available. Second, we trained our system on these seen domains but tested on 5 unseen domains, that only their test data was available. And third, we trained our system on 10 seen domains and tested on all 15 domains. The objective evaluation results for different $B$ values are shown in Table \ref{table15} and the results of all comparisons are shown in Table \ref{table16},\ref{table17} and \ref{table18}. As can be seen,  in $B=1$ which is actually a greedy search, the scores are low and by using larger values for $B$ the scores are also increased. In the experiment on unseen domains, since many meaning labels were entirely specific to these domains and were not shared with the seen domains, our system did not outperform the best-submitted system. But for seen domains and overall domains, our system achieved a comparable result against other submitted methods. Table \ref{table19} shows the mean scores of each human evaluation factor. As can be seen, the sentences generated by our system received high scores from the judges for the informativeness, naturalness and quality factors.

\begin{table}
\begin{center}
\captionsetup{%
    font={small}, 
    format=plain,  
    singlelinecheck=true,      
    labelsep=newline,
    labelfont=bf,      
}
\caption{Performance of the proposed system on WebNLG dataset based on different Beam widths (B) and number of candidate sentences (C) in terms of the BLEU, METEOR and TER scores.}
\label{table15}
\fontsize{8.5}{15}\selectfont
\begin{tabular}{+l ^c ^c ^c^c^c}
\hline
\rowstyle{\bfseries}
Metric & B=1, C=1 & B=5, C=5 & B=10, C=10 & B=15, C=15 & B=20, C=20\\ 
 \hline
BLEU & 37.05 & 40.71 & 43.78 & 46.13 & 47.83\\
METEOR & 0.3611 & 36.87 & 0.3721 & 0.3803& 0.3869\\
TER& 0.62 & 0.56 & 0.52 & 0.47 & 0.45\\
 \hline
\end{tabular}
\end{center}
\end{table}
\begin{table}
\begin{center}
\captionsetup{%
    font={small}, 
    format=plain,  
    singlelinecheck=true,      
    labelsep=newline,
    labelfont=bf,      
}
\caption{Performance of different models on WebNLG dataset for seen domains in terms of the BLEU, METEOR and TER metrics; bold denotes the best scores\textsuperscript{*}.}
\label{table16}
\fontsize{7.5}{15}\selectfont
\begin{tabular}{>{\centering\arraybackslash}{l} >{\centering\arraybackslash}{c} >{\centering\arraybackslash}{c} >{\centering\arraybackslash}{c}}
\hline
\rowstyle{\bfseries}
 Methods &  BLEU  &  METEOR & TER  \\ \hline
Baseline & 52.39 & 37.00 & 0.44 \\
Melbourne & 54.52 & 41.00 & 0.40 \\
TILB-SMT & 54.29 & 42.00 & 0.47 \\
PKUWRITER & 51.23 & 37.00 & 0.45 \\
UPF-FORGE & 40.88 & 	40.00 & 0.55\\ 
Our system & \textbf{60.74} & \textbf{45.80} & \textbf{0.36}\\ 
 \hline
  \multicolumn{4}{p{0.5\textwidth}}{\textsuperscript{*}\small \textit{Notes. The scores for baseline and other systems are reported scores by their authors.}} 
\end{tabular}
\end{center}
\end{table}
\begin{table}
\begin{center}
\captionsetup{%
    font={small}, 
    format=plain,  
    singlelinecheck=true,      
    labelsep=newline,
    labelfont=bf,      
}
\caption{Performance of different models on WebNLG dataset for unseen domains in terms of the BLEU, METEOR and TER metrics; bold denotes the best scores\textsuperscript{*}.}
\label{table17}
\fontsize{7.5}{15}\selectfont
\begin{tabular}{>{\centering\arraybackslash}{l} >{\centering\arraybackslash}{c} >{\centering\arraybackslash}{c} >{\centering\arraybackslash}{c}}
\hline
\rowstyle{\bfseries}
 Methods &  BLEU  &  METEOR & TER  \\ \hline
Baseline & 6.13	& 0.07 & 0.80 \\
Melbourne & 33.27 & 33.00 & 0.55 \\
TILB-SMT & 29.88 & 33.00 & 0.61 \\
PKUWRITER & 25.36	 & 24.00 & 0.67 \\
UPF-FORGE & \textbf{35.70}	 & \textbf{37.00} & 0.55\\ 
Our system & 33.80 & 34.07 & \textbf{0.51}\\ 
 \hline
  \multicolumn{4}{p{0.5\textwidth}}{\textsuperscript{*}\small \textit{Notes. The scores for baseline and other systems are reported scores by their authors.}} 
\end{tabular}
\end{center}
\end{table}
\begin{table}
\begin{center}
\captionsetup{%
    font={small}, 
    format=plain,  
    singlelinecheck=true,      
    labelsep=newline,
    labelfont=bf,      
}
\caption{Performance of different models on WebNLG dataset for all domains in terms of the BLEU, METEOR and TER metrics; bold denotes the best scores\textsuperscript{*}.}
\label{table18}
\fontsize{7.5}{15}\selectfont
\begin{tabular}{>{\centering\arraybackslash}{l} >{\centering\arraybackslash}{c} >{\centering\arraybackslash}{c} >{\centering\arraybackslash}{c}}
\hline
\rowstyle{\bfseries}
 Methods &  BLEU  &  METEOR & TER  \\ \hline
Baseline & 33.24& 23.00 & 0.61 \\
Melbourne & 45.13 & 37.00 & 0.47 \\
TILB-SMT & 44.28 & 34.00 & 0.53 \\
PKUWRITER & 39.88 & 31.00 & 0.55 \\
UPF-FORGE & 38.65	 & \textbf{39.00} & 0.55\\ 
Our system & \textbf{47.83} & 38.69 & \textbf{0.45}\\ 
 \hline
  \multicolumn{4}{p{0.5\textwidth}}{\textsuperscript{*}\small \textit{Notes. The scores for baseline and other systems are reported scores by their authors.}} 
\end{tabular}
\end{center}
\end{table}
\begin{table}
\begin{center}
\captionsetup{%
    font={small}, 
    format=plain,  
    singlelinecheck=true,      
    labelsep=newline,
    labelfont=bf,      
}
\caption{Results of Human Evaluations WebNLG datasets in terms of Informativeness, Naturalness and Quality, rating out of 5. For baselines, the reported sentences by challenge organizers are used.}
\label{table19}
\fontsize{7.5}{15}\selectfont
\begin{tabular}{>{\centering\arraybackslash}{l} >{\centering\arraybackslash}{c} >{\centering\arraybackslash}{c} >{\centering\arraybackslash}{c}}
\hline
\rowstyle{\bfseries}
 Methods &  Informativeness  &  Naturalness & Quality  \\ \hline
Baseline & 5.00 & 4.06 & 4.47 \\
Melbourne & 4.98 & 4.11 & 4.51 \\
TILB-SMT & 4.95 & 4.23 & 4.43 \\
PKUWRITER & 4.95 & 4.27 & 4.54 \\
UPF-FORGE & 4.93	 & 4.27 &4.67\\ 
Our system & 5.00 & 4.63 & 4.67\\ 
 \hline
\end{tabular}
\end{center}
\end{table}

\section{Conclusion}
We presented a new stochastic corpus-based approach for Natural Language Generation using dependency information for sentence structuring and surface realization. At the training time, the proposed model encodes the dependency relations between words of training utterances through a set of features and at the test time, the corresponding dependency tree for a given meaning representation is produced by concatenating the extracted dependency features. The final sentences are generated from the produced dependency tree. We assessed our proposed system on nine different NLG domains of both tabular and dialogue act formats, also datasets of E2E and WebNLG challenges, through the individual domain, general domain, and unseen domain experiments. By comparing our model against the state-of-the-art NLG models, for the individual domain and general domain experiments, the proposed model empirically shows consistent improvement over the statistic data-to-text methods, that are trained on tabular datasets, and also achieves comparable results with the neural network-based models that trained on dialogue act, E2E and WebNLG datasets. Furthermore, our proposed model shows an ability of adaptation to an unseen domain. Also, human evaluation results show that the proposed model can generate high-quality and fluent sentences. 

\starttwocolumn
\bibliography{compling_style}
\end{document}